\documentclass{article} %

\usepackage{geometry}
\usepackage{pdflscape}

\usepackage{microtype}
\usepackage{graphicx}
\usepackage{booktabs,siunitx}
\usepackage{multirow}
\usepackage{makecell}
\usepackage{textcomp}

\usepackage{xcolor,pifont}
\newcommand*\colourcheck[1]{%
  \expandafter\newcommand\csname #1check\endcsname{\textcolor{#1}{\ding{52}}}%
}
\newcommand*\colourx[1]{%
  \expandafter\newcommand\csname #1xmark\endcsname{\textcolor{#1}{\ding{55}}}%
}
\colourcheck{green}
\colourx{red}
\colourx{black}

\newcommand{\mc}[3]{\multicolumn{#1}{#2}{#3}}

\usepackage{hyperref}
\usepackage{appendix}

\usepackage{natbib}
\usepackage[accepted]{icml2023}
\usepackage{color,soul}

\usepackage{nicefrac}
\usepackage{amsmath}
\usepackage{amsthm}
\usepackage{subcaption}
\usepackage{lipsum} %

\usepackage{catchfile}
\usepackage{kotex}
\usepackage[greek,english]{babel}

\renewcommand{\textapprox}{\raisebox{0.5ex}{\texttildelow}}

\newcommand{\modelparams}{g_w}
\newcommand{\sdd}{SemDeDup}

\DeclareTextFontCommand{\textarmenian}{\armenian}

\usepackage{xstring}
\usepackage{textcomp}

\usepackage{bbold}
\usepackage{amssymb}
\usepackage{lipsum}                     %
\usepackage{xargs}                      %
\usepackage{upquote}
\usepackage[OT6, T1]{fontenc}  
\usepackage[utf8]{inputenc}
\newcommand{\straightquote}{\textquotedbl{}}

\usepackage{contour}
\usepackage[normalem]{ulem}

\contourlength{0.8pt}

\newcommand{\myuline}[1]{%
  \uline{\phantom{#1}}%
  \llap{\contour{white}{#1}}%
}

\RequirePackage{xparse}
\ExplSyntaxOn
\NewDocumentCommand{\myulineX}{m}
 {
  \seq_set_split:Nnn \l_tmpa_seq { ~ } { #1 }
  \seq_map_inline:Nn \l_tmpa_seq { \myuline{##1} ~ } \unskip
 }
\ExplSyntaxOff

\newcommand{\uliner}[1]{
  \myulineX{#1}
}

\usepackage[colorinlistoftodos,textsize=tiny,textwidth=2.1cm,shadow,loadshadowlibrary]{todonotes}
\newcommandx{\unsure}[2][1=]{\todo[linecolor=red,backgroundcolor=red!25,bordercolor=red,#1]{#2}}
\newcommandx{\change}[2][1=]{\todo[linecolor=blue,backgroundcolor=blue!25,bordercolor=blue,#1]{#2}}
\newcommandx{\info}[2][1=]{\todo[linecolor=OliveGreen,backgroundcolor=OliveGreen!25,bordercolor=OliveGreen,#1]{#2}}
\newcommandx{\improvement}[2][1=]{\todo[linecolor=Plum,backgroundcolor=Plum!25,bordercolor=Plum,#1]{#2}}
\newcommandx{\thiswillnotshow}[2][1=]{\todo[disable,#1]{#2}}
\newcommandx{\aleks}[1]{\todo{Aleks: #1}}
\newcommandx{\logan}[1]{\todo[linecolor=blue,backgroundcolor=blue!25,bordercolor=black,shadow]{Logan:
#1}}
\newcommandx{\axel}[1]{\todo[linecolor=orange,backgroundcolor=yellow,bordercolor=black]{Axel:
#1}}

\usepackage{amsmath,amsfonts,bm}

\def\1{\bm{1}}

\DeclareMathAlphabet{\mathsfit}{\encodingdefault}{\sfdefault}{m}{sl}
\SetMathAlphabet{\mathsfit}{bold}{\encodingdefault}{\sfdefault}{bx}{n}

\DeclareMathOperator*{\argmin}{arg\,min}

\usepackage{url}
\usepackage[shortlabels]{enumitem}

\usepackage{bm}
\usepackage{mathtools}
\usepackage{amsmath}

\newcommand{\E}{\mathbb{E}}

\newcommand{\dsdm}{\textsc{DsDm}}
\newcommand{\dsir}{\textsc{DSIR}}
\newcommand{\heur}{\textsc{Classifier}}
\newcommand{\randommethod}{\textsc{Random}}

\newcommand{\dtarg}{\mathcal{D}_{\mathrm{targ}}}
\newcommand{\twox}{2\texttimes{}}

\newcommand{\ALog}{\mathcal{A}_\mathrm{Log}}

\newcommand{\ind}[1]{\text{\usefont{U}{bbold}{m}{n}1}_{ #1 }}
\newcommand{\defeq}{\vcentcolon=}

\newcommand{\hatLdm}{\widehat{\mathcal{L}}_{\dtarg{}}}
\newcommand{\Sdm}{\widehat{S}_{\mathrm{DM}}}
\newcommand{\argtopk}{\mathop{\mathrm{arg\,bot}\mbox{\hspace{0.6pt}-}k}}

\newcommand{\lambada}{LAMBADA}
\newcommand{\csalg}{CS-Algorithms}
\newcommand{\squad}{SQuAD}
\newcommand{\jeopardy}{Jeopardy}

\author{
    Logan Engstrom \\
    \texttt{engstrom@mit.edu} \\
    MIT
    \and
    Axel Feldmann \\
    \texttt{axelf@mit.edu} \\
    MIT
    \and
    Aleksander M\k{a}dry \\
    \texttt{madry@mit.edu}\\
    MIT
}

\date{}

\title{\dsdm{}: Model-Aware Dataset Selection with Datamodels}

\begin{document}

\maketitle

\begin{abstract}
  When selecting data for training large-scale models, standard practice is to
filter for examples that match human notions of data quality. Such filtering
yields qualitatively clean datapoints that intuitively should improve model
behavior. However, in practice the opposite can often happen: we find that
selecting according to similarity with ``high quality'' data sources may not
increase (and can even \textit{hurt}) performance compared to randomly selecting
data.

To develop better methods for selecting data, we start by framing dataset
selection as an optimization problem that we can directly solve for: given
target tasks, a learning algorithm, and candidate data, select the subset that
maximizes model performance. This framework thus avoids handpicked notions of
data quality, and instead models explicitly how the learning process uses train
datapoints to predict on the target tasks. Our resulting method greatly improves
language model (LM) performance on both pre-specified tasks and
\textit{previously unseen} tasks. Specifically, choosing target tasks
representative of standard LM problems and evaluating on diverse held-out
benchmarks, our selected datasets 
provide a \twox{} compute multiplier over baseline methods.

\end{abstract}

\section{Introduction}
\label{sec:intro}
Suppose we want to train a large-scale machine learning model. What data should
we train on? The simple answer is: as much data as possible. For example, we
train language and vision models on vast quantities of
text~\citep{radford2019language} and image-caption~\citep{ramesh2021zero} data
from sources like internet crawls. This seemingly straightforward recipe yields
models that generalize remarkably well to a broad range of tasks.

A closer look, however, reveals that choosing training data is not actually so
straightforward. Indeed, not all data is equally useful; for example, internet
data sources frequently contain ``low quality'' data like
spam, poor writing, or nonsense text. Therefore, in practice, we tend to filter
training data according to intuitive notions of quality, e.g., choosing
documents similar to a ``high quality'' data source like Wikipedia or discarding
documents with fewer than five sentences. These steps choose (qualitatively)
``clean'' samples that should \emph{intuitively} improve performance. %
However, do such samples improve performance in practice too?

\textbf{Contributions.} We find that the opposite can happen: selecting data
according to similarity with ``high quality'' data sources may not improve (and,
in fact, can even hurt) model performance. Specifically, we train language
models with standard, similarity-based selection methods previously used to
select data for models like PaLM and
GPT-3~\citep{brown2020language,xie2023data}, and find these methods do not
outperform (and can even underperform) selecting data at random (cf.
Section~\ref{sec:scaling}).

To develop better methods for selecting training
data, we start from first principles. That is, we avoid intuitive notions of
data quality, and instead frame dataset selection as an optimization problem
where the goal is to---given target tasks, a learning algorithm, and a candidate
data pool---select the data that maximizes model performance. However, actually
finding the optimal solution to this problem is difficult. While we can
calculate the performance of a \emph{specific} training set by training a model
on that set (and then evaluating), it is (generally) unclear how to calculate
the \textit{best} possible training subset without examining every possible
subset one by one, a computationally infeasible procedure.

We instead \textit{approximate} the optimal subset by (approximately) modeling
how the learning algorithm actually uses training data to predict. %
Specifically, in Section~\ref{sec:methods}, we model target task performance as
a function of training subset using datamodels (which efficiently approximate
the mapping between training subset and model
performance~\citep{ilyas2022datamodels}), and select the subset that maximizes
our estimate. Then, in Section~\ref{sec:science}, we demonstrate that our
resulting method, \textit{dataset selection with datamodels} (\dsdm{}),
consistently improves language model performance on diverse target tasks (e.g.,
\squad{}~\citep{rajpurkar2016squad} and \lambada{}~\citep{paperno2016lambada}),
even when existing selection methods do not.

\dsdm{}-selected data can improve performance on pre-specified tasks. However,
in practice we train large-scale models to generalize to \textit{yet unseen}
tasks. Our framework suggests a principled approach to selecting data in this
scenario too: choose target tasks similar to those we expect at deployment time,
then select the optimal dataset subset for these target tasks. Following this
strategy, in Section~\ref{sec:scaling}, we choose target tasks that cover a
range of natural language problem categories (\squad{},
\jeopardy{}~\citep{mosaicml2023llm}, and \lambada{}), and select data from C4, a
canonical web crawl~\citep{raffel2020exploring}. Our selections deliver a
\textit{\twox{} compute multiplier} on a diverse set of test benchmarks:
\dsdm{}-selected datasets yield LMs that perform as well as those trained with
\twox{} the compute budget on randomly selected data (we train up to 1.8B
parameter models). 
In contrast, no baseline method outperforms randomly selecting data---even at
the same compute budget.

\section{Estimating the optimal dataset selection with \dsdm{}}
\label{sec:methods}
To select better data for training large-scale models, we start by defining the
optimal dataset selection as an optimization problem. We then select data by
finding a train subset that is \emph{approximately} the best solution to that
problem. Specifically, we use datamodels~\citep{ilyas2022datamodels} to
approximate how the learning algorithm uses data to predict on the tasks of
interest. We describe the resulting framework in more detail below.

\subsection{Task-optimal dataset selection}
We frame dataset selection as an optimization problem where the goal is to
minimize trained model loss on a set of target tasks with respect to training
data choice. Given a learning algorithm $\mathcal{A}$ (e.g., SGD on a neural
network) that maps train set to trained model, and a target distribution
$\dtarg{}$ (e.g., a language modeling task), the size-$k$ \textit{task-optimal
dataset selection} over the set $\mathcal{S}$ of available data (e.g., documents
from an internet scrape) is the subset
\begin{align}
    S^* &\defeq{} \argmin\limits_{S \subset \mathcal{S}, |S| = k} \mathcal{L}_{\dtarg{}}(S), \label{eq:ods}\\
    \mbox{where } \mathcal{L}_{\mathcal{D}}(S) &\defeq{} \E_{x \sim\mathcal{D}} \nonumber
     \left[\ell(x; \mathcal{A}(S))\right],
\end{align}
that minimizes the trained model population loss $\mathcal{L}_{\dtarg{}}(S)$, where
$\ell(x; g)$ denotes the loss (e.g., cross-entropy
loss) for model $g$ on example $x$. Note the expectation in the population loss is over both target dataset
\textit{and} learning algorithm randomness (as, e.g., SGD is a non-deterministic
algorithm). 

In our setting, minimizing \eqref{eq:ods} is difficult. Indeed, we do not have
an easy-to-optimize, closed-form expression for trained model loss in terms of
training set choice $S$ for large-scale model learning
algorithms.
\footnote{Depending on the setup, we may have such a form for other
classes of learning algorithms, like linear regression (with influence
functions~\citep{cook1977detection,giordano2019swiss}) or kernel
regression~\citep{bierens1988nadaraya}.} 
While we can directly calculate the
trained model loss for a given $S$ by actually training on $S$ with
$\mathcal{A}$ (and then evaluating loss), using this method to find the best
subset is generally computationally infeasible:
we would need to train (and evaluate) a model for each of the ${|\mathcal{S}|
\choose k}$ possible size-$k$ train subsets.

\subsection{Estimating model loss efficiently with datamodels}
\label{subsec:dms}
To circumvent this computational challenge, we trade optimality for feasibility,
and instead \textit{estimate} the best train subset.
Specifically, we \textit{approximate} the trained model loss in place of calculating
it directly, then \textit{select} the subset that minimizes our approximation.

The core primitive we use to approximate the trained model loss is
datamodeling~\citep{ilyas2022datamodels}, a framework originally designed to
predict how choice of training set changes model predictions. More precisely, a
datamodel for a fixed sample $x$ approximates the mapping from train subset
choice $S$ (out of the available dataset $\mathcal{S}$) to resulting trained
model loss on a sample $x$, i.e., the function:
$$\mathcal{L}_x(S) \defeq \E\left[\ell(x; \mathcal{A}(S))\right].$$ 

Previous work used datamodels primarily for reliability purposes, e.g., to
detect data poisoning~\citep{khaddaj2022backdoor} or train-test
leakage~\citep{ilyas2022datamodels}. In contrast, we leverage datamodels to
cheaply approximate the trained model loss $\mathcal{L}_x$.
Formally, given a candidate data subset $S\subset\mathcal{S}$, datamodels
take as input the corresponding characteristic vector
\begin{equation}
    \label{eq:characteristic}
    \ind{S} \in \{0, 1\}^{\vert \mathcal{S} \vert}\hspace{10pt}\hbox{ such that}\hspace{10pt}\left(\ind{S}\right)_i = \begin{cases} 
        1 & \text{if } \mathcal{S}_i \in S \\
        0 & \text{otherwise}
        \end{cases},
\end{equation}
instead of the subset $S$ directly. 
Then, the datamodel $\tau_{\theta_x}$ for $x$ is the parameterized function that
optimally
predicts
$\mathcal{L}_x$ over a (chosen) distribution of train subsets
$\mathcal{D}_\mathcal{S}$, i.e.,
\begin{align}
    \label{eq:dm_obj}
    \tau_{\theta_x}: \{0,1\}^{|\mathcal{S}|} \to \mathbb{R}, \qquad\text{ where }\qquad
    \theta_x =
    \arg\min_{\theta}\ \widehat{\E}^{(m)}_{S_i \sim \mathcal{D}_\mathcal{S}}
    \left[ L_{\normalfont{\text{reg}}}\left(
        \tau_\theta(\ind{S_i}),\ \mathcal{L}_x(S_i)
        \right) \right],
\end{align}
where $L_{\normalfont{\text{reg}}}(\cdot, \cdot)$ is a regression loss function
(e.g., mean squared error), and $\widehat{\mathbb{E}}^{(m)}$ is an $m$-sample
empirical expectation. Note that in practice, we \textit{estimate} the
datamodel parameters that minimize \eqref{eq:dm_obj} (i.e., we estimate
the parameters of the function we use to approximate model loss).

\paragraph{Linear datamodels.}
So far we have only defined the datamodeling framework; we have not actually
defined the parameterized function $\tau_\theta$ or described how to estimate  the parameters $\theta$. In
this work, we instantiate datamodels as a \textit{linear} function of the characteristic vector $\ind{S}$ (a standard
choice~\citep{ilyas2022datamodels,saunshi2023understanding}), such that
\begin{equation*}
    \tau_{\theta_x}(\ind{S}) \defeq \theta_x^\top\ind{S}.
\end{equation*}
Note that, being a linear model, $\tau_{\theta_x}$
treats the inclusion of an example $\mathcal{S}_i$ in the train set as having a
fixed effect on $\mathcal{L}_x(S)$ irrespective of the other examples in $S$
(this fixed effect is exactly the value of index $i$ of $\theta_x$).

In this work, to estimate linear datamodel parameters $\theta_x$ we largely
follow the procedures of previous
work~\citep{park2023trak,ilyas2022datamodels}---in particular, we use the TRAK
estimator---but make changes needed for the language modeling domain (see
Appendix~\ref{app:datamodel_estimation} for full details).

\subsection{\dsdm{}: Dataset Selection with Datamodels} Recall that our goal is
to estimate the candidate data subset that minimizes trained model loss on the
target task (cf. \eqref{eq:ods}). To do so, we approximate the mapping between
training subset $S$ and target distribution loss $\left(\textrm{i.e.,
}\mathcal{L}_{\dtarg}(S)\right)$ with datamodels as a primitive, then select the
candidate data subset that minimizes our approximation of the target loss.

Specifically, given a train subset $S$, we estimate the corresponding target
distribution loss with an $n$-sample empirical expectation of datamodel loss
estimates over $\dtarg{}$ samples:
\begin{equation*}
\hatLdm{}(S) = \widehat{\E{}}^{(n)}_{x_i \sim \dtarg{}} \left[\tau_{\theta_{x_i}}(\ind{S})\right] = \frac{1}{n} \sum_{i=1}^n \theta_{x_i}^\top \ind{S} = \ind{S}^\top
\left(\frac{1}{n} \sum_{i=1}^n \theta_{x_i}\right).
\end{equation*}
Then, our size-$k$ \textit{dataset selection with datamodels} (\dsdm{}) estimate
of the optimal dataset selection is the subset that minimizes the approximated
target loss $\hatLdm{}(S)$ with respect to training set choice:
\begin{equation*}
    \Sdm \defeq \argmin_{S \subset \mathcal{S}, |S|=k} \hatLdm{}(S) = \argmin_{S \subset \mathcal{S}, |S|=k} \ind{S}^\top
    \left(\frac{1}{n} \sum_{i=1}^n \theta_{x_i}\right) = \argtopk{}\left(\frac{1}{n} \sum_{i=1}^n \theta_{x_i}\right).
\end{equation*}
In our instantiation, the considered datamodels are linear, so \dsdm{} selects
the examples corresponding to the smallest $k$ indices of $\frac{1}{n}
\sum_{i=1}^n \theta_{x_i}$. (Note that linear datamodels are a design choice:
\dsdm{} can use any datamodel parameterization that can be optimized over.)

\section{Evaluating \dsdm{}}
\label{sec:science}
To what extent does \dsdm{} actually minimize trained model target task loss? In
this section, we demonstrate that \dsdm{} consistently reduces LM target task
loss in practice. In contrast, baseline targeted dataset selection methods---all
of which ignore the model training process and instead select data according to
textual similarity with target task samples---often do \textit{not} outperform
randomly selecting training data. Below, we describe our experimental setup,
then discuss results.

\subsection{Setup}
To capture the effectiveness of a given data selection method, we measure the extent to which it reduces the
optimal dataset selection objective of \eqref{eq:ods},
\begin{equation*}
    \mathcal{L}_{\dtarg{}}(S) \defeq \E_{x \sim\mathcal{D}}\left[\ell(x; \mathcal{A}(S))\right],
\end{equation*}
across varying target tasks. For each considered target task, we split samples
into a \textit{target} set and a separate \textit{test} set, and only use the
target set to select training subsets. We then train an LM on the resulting
dataset, and inspect target task performance (using the test set). Below, we
describe the experimental setup as well as the baselines we use (see
Appendix~\ref{app:science} for more setup details).

\paragraph{Target tasks, candidate dataset, and model training.}
We consider four separate LM target tasks:
\lambada{}~\citep{paperno2016lambada}, \csalg{}~\citep{srivastava2022beyond},
\squad{}~\citep{rajpurkar2016squad}, and
\jeopardy{}~\citep{tunguz2019jeopardy!}; see Appendix~\ref{app:science_tasks}
for more details on each task. Our candidate dataset $\mathcal{S}$ is the
English subset of the Colossal Cleaned Common Crawl (C4), a standard web
scrape~\citep{raffel2020exploring}.\footnote{Each candidate example
$\mathcal{S}_i$ is a sequence-length (1024 token) corpus slice; $|\mathcal{S}|
\approx 217{,}000{,}000$ (cf. Appendix~\ref{app:candidate_dataset}).}
On each selected train dataset, we train a 125M parameter GPT-2 style model on
6 billion tokens.

\paragraph{Baselines.}
We compare \dsdm{} with two standard targeted dataset selection methods, both
of which select according to textual similarity between candidate training
samples and $\dtarg{}$ samples: \heur{} (selects the top examples in
$\mathcal{S}$ given by a logistic model trained to classify, on FastText
features, between $\mathcal{S}$ and $\dtarg{}$ samples; used by GPT-3/PaLM/The
Pile~\citep{chowdhery2022palm,gao2020pile}) and \dsir{} (Data
Selection with Importance Resampling chooses train samples with n-grams that
distributionally match those of $\dtarg{}$~\citep{xie2023data}). We also compare
with randomly selecting data (\randommethod{}).

\begin{figure}[tpb]
    \centering
    \includegraphics[width=0.9\textwidth]{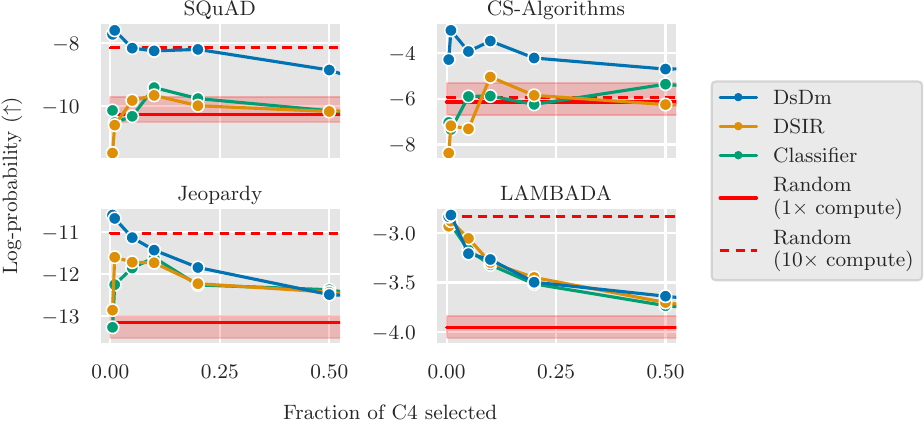}
    \caption{Target task performance by selection method, varying dataset
    selection size. We train a 125M models on a fixed number of tokens for each
    selection, adjusting epochs accordingly. \dsdm{} consistently improves
    performance, even when baselines do not outperform randomly selecting data
    (e.g., on \squad{} and \csalg{}). \dsdm{} models also consistently match a
    larger model trained with 10\texttimes{} the compute budget on random data
    (a Chinchilla-optimal 1.3B model). \dsdm{} performance decreases with larger
    selection fraction, indicating that higher ranked \dsdm{} samples (i.e.,
    data in the smallest selections) tend to improve performance more than less
    highly ranked samples (i.e., data only present in larger selections). We
    measure the average log-probability of the label across samples. The
    ``random'' shaded area is the range of values achieved by 10 \randommethod{}
    models trained on one epoch of data (\randommethod{} performance is not
    x-axis dependent). Measuring accuracy in place of log-probability yields
    similar conclusions (cf. Figure~\ref{fig:accuracy_iid}).}
    \label{fig:dsdm_iid}
\end{figure}

\subsection{Results}
In Figure~\ref{fig:dsdm_iid} we display the mean log-probability (of the label
given the context, across task samples; larger is better) achieved on each
target task by training a model with each selection method (varying dataset
selection size). Each model was trained on the same number of total tokens, with
models trained on smaller fractions of C4 traversing more epochs. We find that
\dsdm{} most improves target task performance on all tasks. Models trained with
\dsdm{} even outperform a larger model trained with 10\texttimes{} the compute
on randomly selected data. Additionally, \dsdm{} performance decreases with
larger selection fraction, indicating that the samples predicted by \dsdm{} to
most improve performance actually do so in practice. After all, smaller
selections will contain more useful data (as predicted by \dsdm{}) on average
compared to larger selections (e.g., all methods select the same subset for
selection fraction $1$).

In contrast, baselines that select according to textual similarity with the
target task, \heur{} and \dsir{}, do \textit{not} consistently outperform
randomly selecting data (e.g., on \squad{} and \csalg{}). These results suggest
that similarity with the target task does \textit{not} suffice to find useful
train samples. Note that the only task that baselines match \dsdm{} on,
\lambada{} (a passage completion task), is also the only one without contextual
instructions. We hypothesize that n-gram similarity may not capture how
instructions define tasks.

\begin{figure}[t]
    \centering
    \begin{subfigure}{0.33\textwidth}
        \centering
        \tiny
        \begin{minipage}{\linewidth}
            \raggedright
            \begin{enumerate}[wide, labelwidth=!, labelindent=0pt, label=\scriptsize{\textbf{(\arabic*)}}]
             \item s, forms, and modification alternative can be overwhelming. So save the time, chance, money, budget, energy, also effort and implement these tips to acquire a obvious concept of what you would like and things you need before you start the quest and think about the right variations and pick right decoration, here are some recommendations and photos on deciding on the best leather sectional sofas toronto.\textcolor{blue}{\textbackslash{}n}The design need to create impact to your sofa. Could it be modern, luxury, minimalist, or traditional? Co
            \item ises; soldier of fortune.\textcolor{blue}{\textbackslash{}n}3. a person who undertakes great commercial risk; speculator.\textcolor{blue}{\textbackslash{}n}4. a person who seeks power, wealth, or social rank by unscrupulous or questionable means: They thought John was an adventurer and after their daughter's money.\textcolor{blue}{\textbackslash{}n}\straightquote{}There can be adventurer souls.\straightquote{}\textcolor{blue}{\textbackslash{}n}\straightquote{}There can be adventurer sirs.\straightquote{}\textcolor{blue}{\textbackslash{}n}\straightquote{}There can be adventurer reflexes.\straightquote{}\textcolor{blue}{\textbackslash{}n}\straightquote{}There can be adventurer realises.\straightquote{}\textcolor{blue}{\textbackslash{}n}\straightquote{}There can be adventurer profiles.\straightquote{}\textcolor{blue}{\textbackslash{}n}\straightquote{}There can be adventurer problems.\straightquote{}\textcolor{blue}{\textbackslash{}n}\straightquote{}There can be adventurer paths.\straightquote{}\textcolor{blue}{\textbackslash{}n}\straightquote{}There can be
            \end{enumerate}
        \end{minipage}
        \caption{\dsdm{} samples}
        \label{fig:subfiga}
    \end{subfigure}
    \hspace{0.005\textwidth} %
    \begin{subfigure}{0.32\textwidth}
        \centering
        \tiny
        \begin{minipage}{\linewidth}
            \raggedright
            \begin{enumerate}[wide, labelwidth=!, labelindent=0pt, label=\scriptsize{\textbf{(\arabic*)}}]
            \item in Alexandria, where it was begun; and the Greek Bible of the Hellenistic Jews and the Catholic Church may rightly be styled the Alexandrian Greek version of the Old Testament.\textcolor{blue}{\textbackslash{}n}In the early days of the Church the Septuagint was widely used among the Jews; as a rule, though there are exceptions, when the Old Testament is quoted in the New Testament it is from the Greek, not the Hebrew Bible that the quotation is made. The early Jewish-Christians and the great majority of the Jews had the same Bible, and Gent
            \item the Central Committee of the Party, that is, by the Politburo, the Orgburo (Organizational Bureau), and the Secretariat. The decisions made were implemented through the Presidium of the Supreme Soviet of the USSR, the Council of People’s Commissars of the USSR, the GKO, and the General Headquarters of the Supreme Command, which had been established on August 8. Strategic direction of the armed forces was carried out by the General Headquarters through its working body, the General Staff. Major questions as
            \end{enumerate}
        \end{minipage}
        \caption{\dsir{} samples}
        \label{fig:subfigc}
    \end{subfigure}
    \hspace{0.005\textwidth} %
    \begin{subfigure}{0.31\textwidth}
        \centering
        \tiny
        \begin{minipage}{\linewidth}
            \raggedright
            \begin{enumerate}[wide, labelwidth=!, labelindent=0pt, label=\scriptsize{\textbf{(\arabic*)}}]
            \item ris and St Gleb, dating from the mid-12th century, was much rebuilt in succeeding periods, before being restored to its original shape in the 20th century. The crowning achievement of Chernigov masters was the exquisite Church of St Paraskeba (Pyatnitskaya), constructed at the turn of the 12th and 13th centuries. This graceful building was seriously damaged in the Second World War; its original medieval outlook was reconstructed. The earliest residential buildings in the downtown date from the late 17th cen
            \item their professional careers.\textcolor{blue}{\textbackslash{}n}Dr Simpson’s first line is classic.\textcolor{blue}{\textbackslash{}n}latest date in the year it’s been that cold in 50 years of record keeping.\textcolor{blue}{\textbackslash{}n}Back in March, 2007, Al Gore told Congress that \straightquote{}the science is settled.\straightquote{}\textcolor{blue}{\textbackslash{}n}science is settled. The Sun revolves around the Earth, not vice versa.\textcolor{blue}{\textbackslash{}n}science,\straightquote{} spent the rest of his life under house arrest.\textcolor{blue}{\textbackslash{}n}\& Tax Bill (its actual name) through the House? Hopefully, some \straightquote{}cooler\straightquote{}\textcolor{blue}{\textbackslash{}n}seem, may have nothing to do with global warming.\textcolor{blue}{\textbackslash{}n}Paul, let me give you a little advice.\textcolor{blue}{\textbackslash{}n}You migh
            \end{enumerate}
        \end{minipage}
        \caption{\heur{} samples}
        \label{fig:subfigb}
    \end{subfigure}
    \caption{Train examples selected by each method for \squad{}. Selected
    \heur{} and \dsir{} train examples are intuitively ``high quality,'' and
    more similar to \squad{} samples (which are Wikipedia excerpts with
    questions) than \dsdm{} examples are. \dsdm{} samples do not match \squad{},
    but \textit{do} contain QA-style text---like (1) left (an advertisement
    posing a question) or (2) left (a dictionary definition).
    Samples randomly chosen from the top 0.01\% that each method is most likely
    to select (cf. Appendix~\ref{app:quantiles}).
    ``\textcolor{blue}{\textbackslash{}n}'' is a newline.}
    \label{fig:pos_compare}
\end{figure}

To better understand how dataset choice relates to performance, we inspect the
datapoints each method is most and least likely to select (for \squad{}: in
Figure~\ref{fig:pos_compare}, for all other targets: in
Appendix~\ref{app:quantiles}). We find that:

\paragraph{Useful data is not necessarily similar to the target task (or
intuitively helpful at all).} Looking at selected data for \squad{} in
Figure~\ref{fig:pos_compare}, \dsir{} and \heur{} select data that is more
qualitatively similar to \squad{} samples (which are Wikipedia excerpts with
questions, cf. Appendix Figure~\ref{fig:squad_samples}) than \dsdm{}. Instead,
\dsdm{} samples often contain question answering-related text that does not
match the \squad{} format; \dsdm{} performance shows that qualitatively similar
data is not necessarily the \textit{best} data. However, helpful data is not
always \textit{intuitively} useful. Indeed, the \dsdm{} examples for \csalg{}
and \jeopardy{} (cf. Appendix Figures~\ref{fig:topbot_cs_algorithms_dsdm} and
\ref{fig:topbot_jeopardy_dsdm}) often contain seemingly nonsense text. Yet,
\dsdm{} yields the best models for these tasks. 

\paragraph{\dsdm{} discards ``mislabeled'' data.} Samples that \dsir{} and
\heur{} are least likely to select are qualitatively different from those of
\dsdm{}. Inspecting Appendix Figure~\ref{fig:neg_compare} for data selected for
\squad{}: least likely samples for all methods are incoherent/malformed, but
those of \dsdm{} also often contain \textit{QA text}. Despite this, such \dsdm{}
samples examples hurt model performance: training on them is worse than
selecting randomly (cf. Appendix Figure~\ref{fig:squad_worst}). We liken these
samples to ``mislabeled'' examples from supervised learning, and conjecture
excluding such data could (in part) explain \dsdm{} performance.

\section{Selecting data for broad model capabilities}
\label{sec:scaling}
So far, we have shown that \dsdm{} consistently reduces loss on pre-specified
target tasks. However, when we train large-scale models in practice our hope is
that they will perform well on \textit{yet unseen tasks} too. Our framework
suggests a straightforward approach to improving this kind of performance:
choose target tasks that match those we expect to see at model deployment time,
then estimate the optimal dataset selection for these ``proxy'' target tasks.

In this section, we demonstrate that this approach to selecting data can greatly
improve held-out task performance compared to baselines. Specifically, we
consider three target tasks that cover a broad range of language modeling
problem categories---\lambada{} (language understanding problems), \squad{}
(reading comprehension problems), and \jeopardy{} (world knowledge
problems)---and estimate the optimal training dataset selection for these tasks
(all together) via \dsdm{}. We then compare models trained on this data with
models trained via existing dataset selection baselines. Overall, evaluating on
a diverse set of held-out benchmarks (meant to model ``yet unseen tasks''), we
find that: (a) randomly selecting data is a surprisingly strong baseline---no
baseline selection method outperforms selecting data at random---and (b) our
approach yields models that match those trained with \twox{} the training
compute on randomly selected data. In particular, models trained with our
approach reliably improve performance on benchmarks that are qualitatively
related to the target tasks. We describe our setup below, and defer additional
details to Appendix~\ref{app:broad}.

\paragraph{Model training, scaling \dsdm{}, selection baselines, and
evaluation.} We train GPT-2 style LMs with varying compute budgets. To train the
best possible model for a given compute budget, we use Chinchilla-optimal
parameter-to-train-tokens ratios~\citep{hoffmann2022training} and train up to
1.8B parameter models. To select with \dsdm{}, we use 125M proxy models: we
calculate \dsdm{} subsets for 125M models, then train on these selections at
each compute budget (instead of computing \dsdm{} separately for each model
class). \dsdm{} cost scales linearly with model size, so this procedure greatly
reduces overhead (cf. Appendix~\ref{app:compcost}).
For baselines, we compare with two methods that select via textual similarity
with a specified ``high quality'' data source (\dsir{} and \heur{}, the
baselines of Section~\ref{sec:science}), a data deduplication method
(\sdd{}~\citep{abbas2023semdedup}), and selecting data randomly. We evaluate on
15 standard benchmarks (cf. Table~\ref{tab:main_tab_acc_new}).

\paragraph{Target tasks.} We execute each targeted dataset selection method
using its originally proposed target task. For \dsdm{}, we apply the
framework described above: we select three target tasks that cover a broad range
of LM problem categories---\lambada{}, \squad{}, and \jeopardy{}---then estimate
the optimal dataset selection for these tasks together (i.e., $\dtarg{}$ as an
equal mix of these tasks). For \heur{} and \dsir{}, we target a replication of
the ``high quality'' target distribution proposed by these methods (a mix of
Wikipedia~\citep{foundation2022english},
Books1~\citep{presser2021bookcorpusopen}, and
OpenWebText~\citep{gokaslan2019openwebtext}, cf. Appendix~\ref{app:ood_targ}).

\begin{figure}[tpb]
    \centering
    \includegraphics[width=\textwidth]{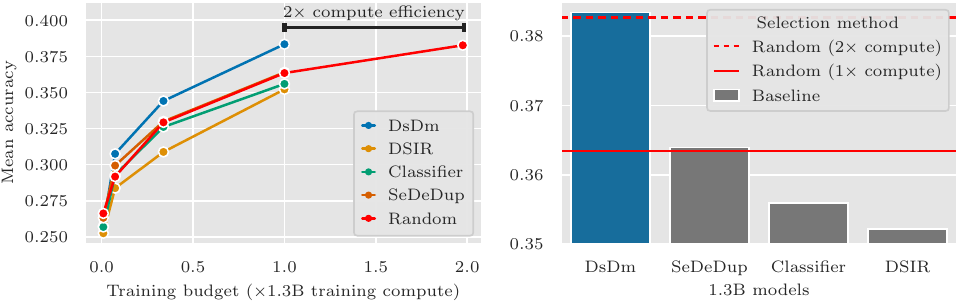}
    \caption{Left: mean benchmark performance, varying training compute budget.
    Right: mean performance for 1.3B models. Randomly selecting data is a strong
    baseline: no active selection baseline outperforms random selection. In
    contrast, \dsdm{} outperforms all baselines across compute budgets (left
    panel), and even matches training with \twox{} the compute on randomly
    selected data (when training 1.3B models, right panel). Our train budgets correspond to training
    125M, 356M, 760M, and 1.3B parameter Chinchilla-optimal LMs. To
    contextualize 1.3B results, we show the performance of a model trained on
    randomly selected data with \twox{} the 1.3B compute budget (i.e., a 1.8B
    Chinchilla-optimal model). For more details, see
    Appendix~\ref{app:broad_model_training_and_target_tasks}.}
    \label{fig:scaling_barplot}
\end{figure}

\subsection{Results}
\label{subsec:maineval}
In Figure~\ref{fig:scaling_barplot}, we display the mean benchmark performance
of models trained with each selection method, varying training compute budget. Randomly
selecting data is a strong baseline: all baseline methods generally match or
perform \textit{worse} than random selection across training compute budgets
(Figure~\ref{fig:scaling_barplot} left). In the case of \heur{} and \dsir{}, we
hypothesize that data selected via similarity with a fixed source hurts model
performance by trading off data diversity for (qualitatively) ``cleaner'' data.

In contrast, \dsdm{} is a \twox{} compute multiplier: \dsdm{}
yields 1.3B models that match models trained with \twox{} the compute
budget on randomly selected data (Figure~\ref{fig:scaling_barplot}, right).
Furthermore, across compute budgets, \dsdm{} consistently outperforms
all selection baselines (Figure~\ref{fig:scaling_barplot}, left). 

\begin{table}[t]
    \centering
    \small
    \caption{Accuracies on the considered benchmarks for 1.3B models trained
    with each selection method, along with a model trained with \twox{} the 1.3B
    compute budget on randomly selected data (a 1.8B model;
    Chinchilla-optimal models with larger parameter counts train with
    more tokens as well). In parentheses, we contextualize
    accuracy with the difference compared to a 1.3B model trained on randomly
    selected data.}
    \label{tab:main_tab_acc_new}
    \begin{tabular}{llllllll}
    \toprule
     &  & \multicolumn{6}{l}{Accuracy (\%)} \\
     & Model Parameters & \multicolumn{5}{l}{1.3B} & 1.8B\\
     & Method & DsDm & Random & Classifier & DSIR & SeDeDup & Random \\
    Category & Benchmark &  &  &  &  &  &  \\
    \midrule
    \multirow[t]{3}{*}{\makecell[tl]{Commonsense\\Reasoning}} & copa & 63.0 (\textcolor[HTML]{006837}{+1}) & 62.0 (\textcolor[HTML]{000000}{+0}) & 66.0 (\textcolor[HTML]{006837}{+4}) & 67.0 (\textcolor[HTML]{006837}{+5}) & 68.0 (\textcolor[HTML]{006837}{+6}) & 64.0 (\textcolor[HTML]{006837}{+2}) \\
     & openbook\_qa & 31.2 (\textcolor[HTML]{A50026}{--2}) & 33.4 (\textcolor[HTML]{000000}{+0}) & 32.0 (\textcolor[HTML]{A50026}{--1}) & 32.0 (\textcolor[HTML]{A50026}{--1}) & 32.2 (\textcolor[HTML]{A50026}{--1}) & 33.6 (\textcolor[HTML]{000000}{+0}) \\
     & piqa & 69.0 (\textcolor[HTML]{000000}{+0}) & 68.9 (\textcolor[HTML]{000000}{+0}) & 69.4 (\textcolor[HTML]{006837}{+1}) & 65.7 (\textcolor[HTML]{A50026}{--3}) & 69.7 (\textcolor[HTML]{006837}{+1}) & 71.5 (\textcolor[HTML]{006837}{+3}) \\
    \midrule
    \multirow[t]{3}{*}{\makecell[tl]{Language\\Understanding}} & cbt & 88.2 (\textcolor[HTML]{006837}{+2}) & 86.4 (\textcolor[HTML]{000000}{+0}) & 85.1 (\textcolor[HTML]{A50026}{--1}) & 92.4 (\textcolor[HTML]{006837}{+6}) & 86.2 (\textcolor[HTML]{000000}{+0}) & 88.4 (\textcolor[HTML]{006837}{+2}) \\
     & hellaswag & 42.3 (\textcolor[HTML]{A50026}{--3}) & 44.9 (\textcolor[HTML]{000000}{+0}) & 42.7 (\textcolor[HTML]{A50026}{--2}) & 40.4 (\textcolor[HTML]{A50026}{--5}) & 44.9 (\textcolor[HTML]{000000}{+0}) & 50.1 (\textcolor[HTML]{006837}{+5}) \\
     & winogrande & 51.1 (\textcolor[HTML]{A50026}{--1}) & 52.2 (\textcolor[HTML]{000000}{+0}) & 50.5 (\textcolor[HTML]{A50026}{--2}) & 55.3 (\textcolor[HTML]{006837}{+3}) & 50.3 (\textcolor[HTML]{A50026}{--2}) & 50.9 (\textcolor[HTML]{A50026}{--1}) \\
    \midrule
    \multirow[t]{3}{*}{\makecell[tl]{Reading\\Comprehension}} & boolq & 58.0 (\textcolor[HTML]{006837}{+3}) & 54.9 (\textcolor[HTML]{000000}{+0}) & 60.9 (\textcolor[HTML]{006837}{+6}) & 61.0 (\textcolor[HTML]{006837}{+6}) & 49.9 (\textcolor[HTML]{A50026}{--5}) & 53.4 (\textcolor[HTML]{A50026}{--2}) \\
     & coqa & 25.5 (\textcolor[HTML]{006837}{+7}) & 18.8 (\textcolor[HTML]{000000}{+0}) & 16.7 (\textcolor[HTML]{A50026}{--2}) & 16.5 (\textcolor[HTML]{A50026}{--2}) & 22.9 (\textcolor[HTML]{006837}{+4}) & 24.9 (\textcolor[HTML]{006837}{+6}) \\
     & news\_qa & 15.6 (\textcolor[HTML]{006837}{+8}) & 7.5 (\textcolor[HTML]{000000}{+0}) & 5.1 (\textcolor[HTML]{A50026}{--2}) & 5.5 (\textcolor[HTML]{A50026}{--2}) & 8.6 (\textcolor[HTML]{006837}{+1}) & 9.5 (\textcolor[HTML]{006837}{+2}) \\
    \midrule
    \multirow[t]{3}{*}{\makecell[tl]{Symbolic\\Problem\\Solving}} & bb\_copy\_logic & 3.1 (\textcolor[HTML]{000000}{+0}) & 3.1 (\textcolor[HTML]{000000}{+0}) & 0.0 (\textcolor[HTML]{A50026}{--3}) & 0.0 (\textcolor[HTML]{A50026}{--3}) & 3.1 (\textcolor[HTML]{000000}{+0}) & 3.1 (\textcolor[HTML]{000000}{+0}) \\
     & bb\_dyck\_lang & 11.9 (\textcolor[HTML]{A50026}{--2}) & 13.5 (\textcolor[HTML]{000000}{+0}) & 3.4 (\textcolor[HTML]{A50026}{--10}) & 1.0 (\textcolor[HTML]{A50026}{--13}) & 7.3 (\textcolor[HTML]{A50026}{--6}) & 8.9 (\textcolor[HTML]{A50026}{--5}) \\
     & bb\_operators & 13.3 (\textcolor[HTML]{006837}{+3}) & 10.5 (\textcolor[HTML]{000000}{+0}) & 6.7 (\textcolor[HTML]{A50026}{--4}) & 10.5 (\textcolor[HTML]{000000}{+0}) & 11.4 (\textcolor[HTML]{006837}{+1}) & 9.5 (\textcolor[HTML]{A50026}{--1}) \\
    \midrule
    \multirow[t]{3}{*}{\makecell[tl]{World\\Knowledge}} & arc\_easy & 47.6 (\textcolor[HTML]{006837}{+3}) & 44.8 (\textcolor[HTML]{000000}{+0}) & 44.7 (\textcolor[HTML]{000000}{+0}) & 39.6 (\textcolor[HTML]{A50026}{--5}) & 43.5 (\textcolor[HTML]{A50026}{--1}) & 48.5 (\textcolor[HTML]{006837}{+4}) \\
     & bb\_qa\_wikidata & 48.1 (\textcolor[HTML]{006837}{+8}) & 40.6 (\textcolor[HTML]{000000}{+0}) & 48.3 (\textcolor[HTML]{006837}{+8}) & 37.7 (\textcolor[HTML]{A50026}{--3}) & 45.5 (\textcolor[HTML]{006837}{+5}) & 53.6 (\textcolor[HTML]{006837}{+13}) \\
     & trivia\_qa & 7.1 (\textcolor[HTML]{006837}{+3}) & 3.7 (\textcolor[HTML]{000000}{+0}) & 2.5 (\textcolor[HTML]{A50026}{--1}) & 3.5 (\textcolor[HTML]{000000}{+0}) & 2.4 (\textcolor[HTML]{A50026}{--1}) & 4.1 (\textcolor[HTML]{000000}{+0}) \\
    \bottomrule
\end{tabular}

\end{table}

Going beyond aggregate performance, we find that \dsdm{} greatly improves on
benchmarks related to the target tasks, while simultaneously not reducing
performance on unrelated categories (on average). More precisely, inspecting
individual benchmark performance in Table~\ref{tab:main_tab_acc_new}, \dsdm{}
most improves reading comprehension and world knowledge benchmarks compared to
selecting randomly. We hypothesize that our choice of target tasks leads to
improved performance on these benchmarks (which are qualitatively similar to
\squad{}, a reading comprehension task, and \jeopardy{}, a world knowledge
task). Furthermore, in these categories \dsdm{} consistently matches or
outperforms training with \twox{} the compute budget on randomly selected data
(i.e., the 1.8B model in Table~\ref{tab:main_tab_acc_new}). Crucially, \dsdm{}
improves on these categories while \textit{also} not reducing performance in
other categories. As a comparison, \dsir{}---which targets mostly formal
text---performs well on language understanding tasks but poorly on other
categories (e.g., world knowledge and symbolic problem solving).

\begin{figure}[t]
    \centering
    \includegraphics{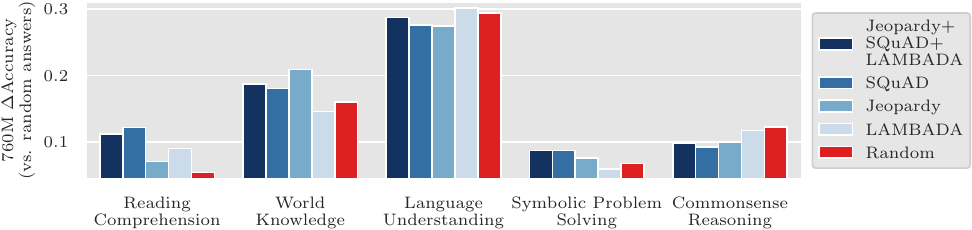}
    \caption{Per-category performance for 760M models trained with
    \dsdm{}-selected data, varying target task. \dsdm{} target tasks generally
    improve performance on (qualitatively) related benchmark task categories.
    Specifically, models targeted towards \squad{}/\jeopardy{}/\lambada{}
    improve accuracy on reading comprehension/world knowledge/language
    understanding, respectively. Targeting all three tasks at once improves
    overall accuracy. However, target tasks can also reduce performance on
    (qualitatively) unrelated tasks. For example, targeting with \lambada{} (a
    language understanding task) reduces performance on world knowledge tasks
    compared to randomly selecting. See each category's constituent benchmarks
    in Table~\ref{tab:main_tab_acc_new}. We plot accuracy improvement compared to
    randomly guessing on benchmarks (e.g., some benchmarks are multiple-choice).}
    \label{fig:varying_targ}
\end{figure}

\paragraph{Target tasks improve performance on qualitatively similar
benchmarks.}
\label{subsec:target_task_role}
So far, we have only targeted \dsdm{} with a mix of \lambada{}, \jeopardy{}
and \squad{}. How does target task choice change model behavior? We find that
targeting a specific task generally improves performance on qualitatively
related tasks. To demonstrate, in Figure~\ref{fig:varying_targ} we display
accuracy by benchmark category while varying target task across \lambada{},
\jeopardy{}, \squad{}, and all at once. Here, targeting a
task generally improves accuracy on related tasks, e.g., \squad{} most
improves reading comprehension, and \jeopardy{} most improves world knowledge.
Furthermore, targeting all tasks at once improves overall accuracy the
most. However, targeting can also decrease accuracy on unrelated tasks. For
example, targeting \lambada{}, a language understanding task, reduces
world knowledge accuracy compared to randomly selecting data. Our
results suggest that we can tailor target tasks to improve deployment-time
performance, but also that we need to be careful to choose targets that are
diverse enough to capture a range of downstream problems.

\paragraph{\dsdm{} is necessary to improve performance (with the targeted
tasks).} \dsdm{} selections yield much better models than \heur{} and \dsir{}
selections. However, we have not yet compared these selection methods
head-to-head with the same \textit{target task}. \heur{} and \dsir{} target a
mix of ``high quality'' sources, while \dsdm{} targets three LM tasks
(\jeopardy{}, \squad{}, and \lambada{}). To what extent does selecting with
\dsdm{} drive performance compared to the difference in target tasks?
We demonstrate that selecting with \dsdm{} is necessary to improve performance
on the considered target tasks. Specifically, we train models on data selected
with \dsir{} and \heur{} targeting \lambada{}, \jeopardy{} and \squad{}, and find
that (just as when targeting ``high quality text'') neither outperforms randomly
selecting data (cf. Appendix Figure~\ref{fig:varying_mech_for_multitask}).

\section{Discussion}
Ostensibly, the sole goal of our dataset selection framework is improve model
performance by better selecting training data. However, one can view our
framework more broadly. That is, one can also use our framework to select data
that boosts any chosen downstream property of our trained models---not just
performance on a given benchmark. In this sense, our framework (and accompanying
method) unlocks data curation as another stage of the model training pipeline
that we can intervene on to control the downstream model behavior in a
fine-grained manner. Below, we discuss in more detail the broader
opportunities this view opens up as well as the other aspects of the framework,
such as proxy modeling and computational efficiency.

\paragraph{Applications and broader opportunities.} \dsdm{} can optimize for any
specified downstream model behavior. Indeed, by an appropriate choice of the
target tasks, we can use our framework to improve a wide range of model
behaviors, including: ``aligning'' models at pretraining time (in addition to or
in place of existing methods, which typically operate post model
training~\citep{bai2022constitutional,ziegler2019fine,taori2023stanford});
optimizing for notions of fairness; and improving performance on specific
domains of interest (such as low-resource languages or programming).

\paragraph{Training stronger models with weaker proxy models.}
We select data for large models by using smaller models to proxy large model
behavior (recall that we use \dsdm{} to select data for smaller proxy models,
then train large models on these selections). Despite that these proxy models
are much worse than larger models on benchmarks (cf. Appendix
Table~\ref{tab:model_training}), the corresponding selections nonetheless
greatly improve performance. Furthermore, training on proxy models' selections is
the simplest possible approach to scaling. Therefore, we suspect that scaling to
larger models less na\"ively could yield even better results. More broadly, our
findings are in line with previous work showing that smaller models can still be
leveraged to determine better training hyperparameters for larger
models~\citep{kaplan2020scaling,coleman2020selection,hoffmann2022training,yang2022tensor,xie2023doremi}.

\paragraph{Computational cost.} 
\dsdm{} is relatively inexpensive to compute in practical model training
scenarios. At a high level, the most expensive part of estimating \dsdm{} is
computing the gradient for each training example on a handful of small proxy
models (in our case, four 125M parameter LMs---see Appendix~\ref{app:compcost}
for a full cost breakdown). To contextualize \dsdm{} cost with model training:
computing gradients also dominates the cost of training LMs. Since the cost of
computing a 125M model gradient is orders of magnitude lower than the cost of
computing gradients for standard model sizes,\footnote{For reference: models
trained today generally range from 3B to 175B parameters. The cost of a gradient
is (roughly) linear in model size, so it is $24\times$ to $1400\times$ more
expensive to compute gradients for these models vs. 125M models.} even a small
compute multiplier (let alone the 2\texttimes{} improvement \dsdm{} seems to
offer) quickly makes the overhead of computing \dsdm{} worthwhile. Additionally,
after computing \dsdm{} on a set of datapoints once, the cost of computing
\dsdm{} on those datapoints again is essentially negligible (as the required
computations are easy to cache). Therefore, we can amortize \dsdm{}'s
computational cost over the entire ``lifetime'' of training on the given
dataset.

\section{Related Work}
Current methods for selecting LM pretraining datasets tend to follow a two-step
framework: (a) choose an intuitively ``high quality'' reference corpus, like
Wikipedia~\citep{foundation2022english}, then (b) select data that matches it.
There are two standard methods that adhere to this framework: \dsir{} (Dataset
Selection with Importance Reweighting~\citep{xie2023data}) and \heur{}
(originally introduced in~\citet{brown2020language} and used by other
work~\citep{gao2020pile,chowdhery2022palm,du2022glam}). Other work on selecting
data for LM pretraining has included deduplicating examples in LM activation
space~\citep{abbas2023semdedup}, and selecting examples with the largest
difference in loss between LMs trained on the candidate and reference
sets~\citep{moore2010intelligent,axelrod2017cynical,feng2022automatic}. Simpler
methods for selecting data are also commonplace. These include removing
documents that are too short or contain too many special
characters~\citep{raffel2020exploring,computer2023redpajama,xie2023data}. In the
LM domain, a related (but different) task to dataset selection is choosing
weights for sampling from mixtures of data
sources~\citep{chen2023skill,xie2023doremi,albalak2023efficient}.

Beyond LM pre-training, previous work also selects data in other domains. These
works aim to: improve the performance on a given
task~\citep{wei2015submodularity,kaushal2019learning,wang2020optimizing,killamsetty2021glister,chitta2021training,mindermann2022prioritized},
identify core-sets of large training
datasets~\citep{sener2017active,phillips2017coresets,coleman2020selection,
mirzasoleiman2020coresets,paul2021deep, killamsetty2021retrieve,
okanovic2023repeated}, and instruction fine-tune
LMs~\citep{chen2023alpagasus,cao2023instruction}. Broadly, such methods select
by prompting pretrained models, discriminating on proxies for model uncertainty
like loss or gradient norm, matching on gradients, or deduplicating in model
output space.

\section{Conclusion}
In this work, we cast dataset selection as an optimization problem: given target
tasks, a learning algorithm, and a candidate training dataset, choose the
training maximizes performance. We then propose a method for approximating the
solution to this optimization problem, \dsdm{}, that selects by modeling how the
learning algorithm uses training data to predict on the target tasks. We show
that our method reliably improves target task performance in the LM setting, and
furthermore use our framework to improve broader model generalization. By
choosing target tasks similar to those we expect to see at deployment time, we
can greatly improve model performance on yet unseen tasks.

Our findings prompt us to take on a much broader view of the role of dataset
selection stage in model training. In particular, our framework demonstrates
that dataset selection can be an effective tool for fine-grain control of model
behavior. Indeed, we hypothesize that carefully choosing data can not only
improve downstream task performance, but also other downstream properties of
trained models, such as notions of predictor fairness, alignment with human
preferences, or capabilities in specific domains like low-resource languages or
programming. We also suspect that current methods for datamodeling only scratch
the surface of understanding how models learn from data---and that we can
greatly improve our ability to manipulate model behavior through training data
by developing better datamodeling techniques.

\subsubsection*{Acknowledgments}
LE funded by the Google PhD Fellowship. Work supported in part by the NSF grants
CNS-1815221 and DMS-2134108, and Open Philanthropy. This material is based upon
work supported by the Defense Advanced Research Projects Agency (DARPA) under
Contract No. HR001120C0015.

\bibliography{bibliography/fmt_bib.bib}
\appendix
\newpage
\appendixpage
\tableofcontents
\newpage
\begingroup
\let\addcontentsline\oldaddcontentsline %
\section{Experimental Setup}
In this section we discuss general experimental setup, including candidate data
pool, considered target tasks, baselines, evaluation metrics, and model training
choices.

\subsection{Candidate dataset}
\label{app:candidate_dataset}
Our candidate dataset is the full English subset of
C4~\citep{raffel2020exploring}. We use the train split of the
\texttt{en.noblocklist} subset of the C4 version prepared by AllenAI at
\url{https://huggingface.co/datasets/c4}. The subset name \texttt{noblocklist}
signifies that curse words were not filtered in the subset.

To split the text from the documents into examples, we tokenize all the
documents, concatenate them together (separated by end-of-text tokens), and then
slice the result into 1024 token chunks. These 1024 token examples generally
contain between 3,000 and 6,000 characters (roughly a thousand words). The final
candidate dataset has 216,948,746 examples. We tokenize with the Pythia
tokenizer~\citep{black2022gpt,biderman2023pythia}.

As a public internet crawl, C4 contains diverse text. To contextualize the
dataset, we show (excerpts) of random C4 samples in
Figure~\ref{fig:random_samples}.

\subsection{Target tasks}
\label{app:target_tasks}
We describe each of the considered target tasks below. We both describe the
tasks, and how we split samples into distinct sets of ``target samples'' (to
select datasets for a target task) and ``holdout samples'' (to evaluate models
on the target task):
\begin{itemize}
\item \textbf{\squad{}.} The Stanford Question-Answering Dataset
(SQuAD~\citep{rajpurkar2016squad}) is an open book, reading comprehension
dataset of questions about Wikipedia articles. The goal is to answer questions
using the corresponding article as context. Our target set is 25\% of the
\squad{} train set (23107 examples), our holdout set is the \squad{} validation
set (10557 examples).
\item \textbf{\jeopardy{}.} Jeopardy~\citep{tunguz2019jeopardy!} is a set of
trivia questions taken directly from the show ``Jeopardy!'' We use the version
of Jeopardy published by MosaicML~\citep{mosaicml2023llm}.\footnote{Located at:
\url{https://github.com/mosaicml/llm-foundry/blob/v0.2.0/scripts/eval/local_data/world_knowledge/jeopardy_all.jsonl}}
We include all the samples save for the ``Word Origins'' subset.\footnote{We
originally intended this subset as a hold-out set for our broader evaluation,
but decided not to use the subset as we deemed it unfairly close to the original
task to serve as a true hold-out set.} We randomly partition the remaining
samples into 876 target samples and 876 holdout samples.
\item \textbf{\lambada{}.} LAnguage Modeling Broadened to Account for Discourse
Aspects (\lambada{}~\citep{paperno2016lambada}) is an open-ended cloze task
measuring broad context text understanding. The goal is to predict the last word
of curated passages from BooksCorpus~\citep{zhu2015aligning} given the rest of
the passage as context. The task is meant to be challenging:
\citet{paperno2016lambada} only select passages such that crowdworkers could not
guess the final word given the final sentence alone (up until the final word),
but could guess the final word given the entire passage. We use the \lambada{}
version curated by EleutherAI.\footnote{Located at
\url{https://huggingface.co/datasets/EleutherAI/lambada_openai/viewer/en}.}
Finally, we split the \lambada{} test set into separate target and holdout sets,
then remove 6 samples from the \lambada{} holdout set due to overlap with
samples in our candidate train dataset (cf. Subsection~\ref{app:ttl} for details
on this procedure). We conclude with 2570 holdout samples and 2577 target
samples.
\item \textbf{\csalg{}.} BIG-bench CS Algorithms~\citep{srivastava2022beyond}
measures the ability of models to solve basic algorithmic problems. In
particular, this benchmark contains two kinds of problems: testing for balanced
parentheses, and finding the longest common subsequence of multiple strings. For
each considered example, the goal is to directly output the answer to the posed
algorithmic question. We randomly split the test set into 660 target samples and
660 holdout samples.
\end{itemize}
We include samples of each benchmark in Figure~\ref{fig:squad_samples}
(\squad{}), Figure~\ref{fig:jeop_samples} (\jeopardy{}),
Figure~\ref{fig:lambada_samples} (\lambada{}), and
Figure~\ref{fig:csalg_samples} (\csalg{}). We evaluate in the 0-shot (for
\lambada{} and \csalg{}) and 3-shot (for \squad{} and \jeopardy{}) regimes. In
the 3-shot setting, we separate each example with a single newline. We use
standard prompts for each task (see the samples for details). 

\clearpage
\begin{figure}[h!]
    \centering
    \begin{minipage}{\linewidth}
                \small
                \raggedright
                \begin{enumerate} %
                \item Context: The chloroplasts of some hornworts and algae
                contain structures called pyrenoids. They are not found in
                higher plants. Pyrenoids are roughly spherical and highly
                refractive bodies which are a site of starch accumulation in
                plants that contain them. They consist of a matrix opaque to
                electrons, surrounded by two hemispherical starch plates. The
                starch is accumulated as the pyrenoids mature. In algae with
                carbon concentrating mechanisms, the enzyme rubisco is found in
                the pyrenoids. Starch can also accumulate around the pyrenoids
                when CO2 is scarce. Pyrenoids can divide to form new pyrenoids,
                or be produced ``de novo''.\\Question: What shape are pyrenoids?\\
                Answer: \hl{roughly spherical}
                \item Context: In this dioxygen, the two oxygen atoms are
                chemically bonded to each other. The bond can be variously
                described based on level of theory, but is reasonably and simply
                described as a covalent double bond that results from the
                filling of molecular orbitals formed from the atomic orbitals of
                the individual oxygen atoms, the filling of which results in a
                bond order of two. More specifically, the double bond is the
                result of sequential, low-to-high energy, or Aufbau, filling of
                orbitals, and the resulting cancellation of contributions from
                the 2s electrons, after sequential filling of the low $\sigma$
                and $\sigma^*$ orbitals; $\sigma$ overlap of the two atomic $2p$
                orbitals that lie along the O-O molecular axis and $\pi$
                overlap of two pairs of atomic $2p$ orbitals perpendicular to
                the O-O molecular axis, and then cancellation of contributions
                from the remaining two of the six 2p electrons after their
                partial filling of the lowest $\pi$ and $\pi^*$ orbitals.\\Question: What
                is a descriptive term for a low-to-high energy bond?\\Answer: \hl{Aufbau}
                \end{enumerate}
    \end{minipage}
    \caption{Random \textbf{\squad{}} samples. Context is normal text, and
    the continuation label is \hl{hightlighted}.}
    \label{fig:squad_samples}
\end{figure}

\begin{figure}[h!]
    \centering
    \begin{minipage}{\linewidth}
                \small
                \raggedright
                \begin{enumerate} %
                \item WORLD HISTORY: In 1191 this Lion-Hearted king of England captured Cyprus \& Acre during the Crusades\\Answer: \hl{Richard I}
                \item LITERATURE: 1719 novel about a mariner who lived 8 \& 20 years all alone in an uninhabited island\\Answer: \hl{Robinson Crusoe}
                \end{enumerate}
    \end{minipage}
    \caption{Random \textbf{\jeopardy{}} samples. Context is normal text, and
    the continuation label is \hl{hightlighted}.}
    \label{fig:jeop_samples}
\end{figure}
        
\begin{figure}[h!]
    \centering
    \begin{minipage}{\linewidth}
                \small
                \raggedright
                \begin{enumerate} %
                \item The Simplification Movement wasn't really an organized
                movement. It was more of an ideological shift by a large number
                of believers. There were quite a few Simpletons among the
                Mother Assembly denomination, but the High Sire had never
                recognized their movement as an order or organization.  However,
                some other denominations were founded on the principles of the
                Simplification \hl{Movement}
                \item ``Here,'' said Jacob, handing them what was a rope
                attached to the ground next to them, the other end at the bottom
                of the well. ``You first.''\\Will stood there. Why am I doing
                this? he thought. \\``Come on, let's go!'' ordered Jacob.\\Will
                took the rope and began to climb down the well.\\``Thatta boy,
                you've got this,'' said \hl{Jacob}
                \end{enumerate}
    \end{minipage}
    \caption{Random \textbf{\lambada{}} samples. We show the context as normal text, and
    the continuation label as \hl{hightlighted}.}
    \label{fig:lambada_samples}
\end{figure}

\begin{figure}[h!]
    \centering
    \begin{minipage}{\linewidth}
        \small
        \raggedright
        \begin{enumerate} %
        \item Given two strings, determine the length of the longest common subsequence.\\~\\Strings: REFVJLZIV PJIQB\\Length of longest common subsequence: \hl{2}
        \item Determine whether the given sequence of parentheses is properly matched.\\~\\Sequence: [ ] ( ) ( ( ( ) ) )\\Valid/Invalid? \hl{Valid}
        \end{enumerate}
    \end{minipage}
    \caption{Random \textbf{\csalg{}} samples. We show the context as normal text, and
    the continuation label as \hl{hightlighted}.}
    \label{fig:csalg_samples}
\end{figure}

\clearpage

\subsubsection{Mitigating train-test leakage}
\label{app:ttl}
We mitigate train-test leakage by filtering out test examples that overlap
with our candidate data samples. Specifically, we define a test example as
``leaked'' if both its context and continuation are present in a single C4
example. To upper-bound train-test leakage, we test for the context and
continuation separately (i.e., for a given test sample, the context and
continuation do not have to be contiguous in a train sample to count as
leaked)
We investigate train-test leakage for all the test examples in each of the test
sets (i.e., \lambada{}, \squad{}, \jeopardy{}, and \csalg{}) across the entire
candidate train set (i.e., the C4 English subset). Note that we match strings
after lowercasing and removing whitespace.

We find 6 \lambada{} test examples with overlap in C4, and remove them from our
\lambada{} test split. We do not find any train-test leakage for \squad{},
\jeopardy{}, or \csalg{}. 

\subsection{Data selection baselines}
\label{app:baselines}
We consider four baselines for selecting language modeling data. These fall into
two categories: \textit{targeted} data selection methods (which select data
according to a target distribution), and \textit{untargeted} data selection
methods (which do not take in a target distribution).

\subsubsection{Targeted baselines}
The two targeted dataset selection methods we consider, \heur{} (originally used
to select the GPT-3 dataset~\citep{brown2020language}) and \dsir{}, both select
according to textual similarity with a target distribution. We describe the
details of these methods below:

\paragraph{\heur{}.} The dataset selection method originally developed to select
data for GPT-3, and additionally used to select data for
PaLM~\cite{chowdhery2022palm} and The Pile~\citep{gao2020pile}. The method
trains a logistic regression model on FastText~\citep{joulin2016bag} features to
classify between (held-out) samples of the candidate dataset (in our case, C4)
and the target distribution, then chooses training data according to how likely
the model predicts the data as being sampled from the target distribution. To
more specifically describe \heur{}: the method keeps a given document if the
scored document satisfies:
$$
\epsilon > 1 - \mathtt{document\_score}, \epsilon \sim \mathrm{Lomax}(\alpha),
$$
where a Lomax sample is drawn for each considered document, and where
$\mathtt{document\_score}$ is the classifier-given probability that the given sample
is in the target distribution. Sampling a threshold according to the Lomax
distribution is meant to improve diversity of the selected data. In this work,
we learn the classifier on the C4 \texttt{en.noblocklist} validation set, and
choose $\alpha=12$ via the parameter selection procedure described in
\citet{brown2020language} (score each document in C4 with the classifier, then
fit the parameters of a Lomax distribution via maximum likelihood estimation
according to these scores).

\paragraph{\dsir{}.} Dataset Selection with Importance
Resampling~\citep{xie2023data} aims to select a data subset with a similar
distribution as the target task in terms of n-gram counts. \dsir{} comprises two
steps: (a) find the (hashed) n-gram counts for each train set example (each
example is represented as a vector of counts, with n-grams hashed into buckets
to reduce dimensionality), then (b) importance sample to select candidate train
set examples that are distributed similarly to target distribution samples in
terms of n-gram counts. \dsir{} calculates importance weights by modeling the
distribution of examples (in feature space) under the target distribution and
under the candidate data distribution separately, using bag-of-words style
models. In greater detail, \dsir{} consists of the following steps:
\begin{enumerate}
    \item Fit $\hat{p}_\mathrm{feat}$ and $\hat{q}_\mathrm{feat}$, estimates of
    the distributions of target examples and candidate training examples in
    hashed n-gram space (respectively). \dsir{} parameterizes
    $\hat{p}_\mathrm{feat}$ and $q_\mathrm{feat}$ through the following general
    procedure for estimating the distribution of hashed n-grams\footnote{For
    example, if we wanted to make a $d$ dimensional hashed n-gram feature vector
    for a document, we would find all the n-grams in the document, hash the
    n-grams into integers up to size $d$, then go through each integer and
    increment the corresponding feature vector index.} for a given set of
    documents. First, calculate the hashed n-gram counts (with $d$ hash buckets)
    across the documents as the vector $\gamma \in \mathbb{R}^d$, where
    $\gamma_k$ corresponds to the number of n-grams that hash to $k$ in the
    documents. Then, normalize $\gamma$ so that its values sum to $1$, forming a
    probability distribution over buckets. Finally, parameterize the
    distribution of hashed n-grams for this set of documents as a bag-of-words
    style model~\citep{zhang2010understanding} such that the probability of a
    document with hashed n-gram counts $c$ is $\prod_{i=1}^d \gamma_d^{c_i}$
    (here, the bag-of-words model is over hashed n-grams instead of words).
    \item Calculate importance weights for each example in the candidate
    training set, such that example $i$ with counts $c$ has weight
    $w_i=\frac{\hat{p}_\mathrm{feat}(c)}{\hat{q}_\mathrm{feat}(c)}$.
    \item Sample examples without replacement according to the categorical
    distribution with (unscaled) weights $w_i$.
\end{enumerate}
For more details on \dsir{}, see Section 4 of \citet{xie2023data}. We adapt
implementations of both \dsir{} and \heur{} from
\url{https://github.com/p-lambda/dsir}. 

\paragraph{Considered target distributions.}
We apply targeted dataset selection methods with different target distributions
depending on the context. In Section~\ref{sec:science}, we measure the extent to
which different selection methods can reduce loss on individual target tasks, so
we select data for \textit{individual tasks} (i.e., \jeopardy{}, \squad{},
\csalg{}, and \lambada{}). In Section~\ref{sec:scaling} we use these targeted
baselines to select data for general purpose language modeling, so we use the
recommended target task from each work (intuitively high-quality data sources;
see Appendix~\ref{app:broad_model_training_and_target_tasks} for more details).

\subsubsection{Untargeted baselines}
The two untargeted dataset selection methods we consider
are: \randommethod{} (select data randomly) and \sdd{} (Semantic
Deduplication~\citep{abbas2023semdedup}). \sdd{} selects by clustering data
according to the last layer activations for the last token in the given
document, then choosing only the examples in each cluster that have the lowest
cosine similarity with the cluster centroid. We follow the hyperparameters from
the original work (11,000 clusters, deduplicating down to 20\% of the dataset
for optimal model performance). We use the implementation from
\url{https://github.com/facebookresearch/SemDeDup/}.

\subsection{LM training details}
\label{app:training_details}
We train GPT-2 family decoder-only transformer
models~\citep{radford2019language,liu2018generating} using
LLM-Foundry~\citep{mosaicml2023llm}. To train models, we use ADAM ($\beta_1=0.9,
\beta_2=0.95,\epsilon=10^{-8}$), sequence length 1024, batch size 1024, a cosine
learning rate schedule (with 200 warm up batches and $\alpha=0.1$), and $\ell_2$
gradient clipping with threshold $1$. We train on A100s (with BF16 precision)
and H100s (with FP8 precision), and tokenize text with the BPE tokenizer used by
Pythia~\citep{biderman2023pythia}.

We summarize the remaining hyperparameter choices used to train the models in
this work in Table~\ref{tab:model_training} (including weight decay, learning
rate, model architecture, and training token count). We select all
hyperparameters to minimize 125M held-out perplexity on C4. The only exception:
we increase the weight decay for the Section~\ref{sec:scaling} models to ensure
that larger parameter model training runs converge (with smaller weight decay,
larger models diverge in loss). Model parameterization choices (i.e., number of
heads or layers), optimizer hyperparameters, and learning rate schedule
generally chosen according to the default LM training configurations in
LLM-Foundry.

\paragraph{Chinchilla-optimal compute ratios.} To train the best possible LM for
a given compute budget, one must trade off two hyperparameters that control used
compute: model size and number of training tokens. We use Chinchilla-optimal
parameter-to-training-token ratios to trade these parameters
off~\citep{hoffmann2022training}. In our compute regime, this (roughly) amounts
to training on a number of tokens equal to 20\texttimes{} the number of
parameters.

\begin{table*}
    \centering
    \caption{Training configurations for models trained across this work.
    Accuracy measured as the mean accuracy across the benchmarks considered in
    Section~\ref{sec:scaling} for a model trained with \textit{randomly
    selected} data with the corresponding configuration. The Chinchilla-optimal
    (760M, 1.3B, 1.8B) models~\citep{hoffmann2022training} of
    Section~\ref{sec:scaling} are much more accurate than the 125M models used
    to calculate datamodels. Following previous work, we approximate FLOPs
    (Floating Point OPerations) via $\texttt{parameters} \times \texttt{tokens}
    \times 6$~\citep{kaplan2020scaling,hoffmann2022training}; FLOPs proxy the
    computational cost of training a given model. LR is learning rate, WD is
    weight decay. Each batch contains 1024 samples of 1024 tokens each.}
    \label{tab:model_training}
    \setlength{\tabcolsep}{0pt}
    \begin{tabular*}{\textwidth}{
      @{\extracolsep{\fill}}
      lcclccllll
    }
    \toprule
    & \mc{6}{c}{Hyperparameters} & & & \\
    \cmidrule{2-7}
    {Parameters} 
    & {LR} & {WD} & {$d_{\textrm{model}}$}
    & {Heads} & {Layers} & {Tokens} & Batches & {Train FLOPs} & \thead{Accuracy} \\
    \midrule
    \mc{9}{@{}l}{\textit{Estimating datamodels}} \\
    125M &  {$6 \times 10^{-4}$} &   {$2 \times 10^{-4}$} &   {$768$} &   {$12$} &  {$12$} & {$8.4 \times 10^{10}$} & $80000$ & {$6.3 \times 10^{19}$} & {$31.8\%$} \\
    \midrule
    \mc{9}{@{}l}{\textit{Section~\ref{sec:science}: Evaluating optimal dataset selection estimators}}\\
    125M &  {$6 \times 10^{-4}$} &   {$2 \times 10^{-4}$} &   {$768$} &   {$12$} &  {$12$} & {$2.6 \times 10^{10}$} & $25000$ & {$2.0 \times 10^{19}$} & {$-$} \\
    \midrule
    \mc{9}{@{}l}{\textit{Section~\ref{sec:scaling}: Evaluating unseen-task generalization (chosen as \textapprox{}Chinchilla-optimal)}}\\
    125M &  {$6 \times 10^{-4}$} &   {$4 \times 10^{-4}$} &   {$768$} &   {$12$} &  {$12$} & {$2.5 \times 10^{9}$}& $2400$ & {$1.9 \times 10^{18}$} & {$26.6\%$} \\
    356M &  {$6 \times 10^{-4}$} &   {$4 \times 10^{-4}$} &   {$1024$} &   {$16$} &  {$24$} & {$7.0 \times 10^{9}$}& $6700$ & {$1.5 \times 10^{19}$} & {$29.2\%$} \\
    760M &  {$6 \times 10^{-4}$} &   {$4 \times 10^{-4}$} &   {$1536$} &   {$12$} &  {$24$} & {$1.5 \times 10^{10}$}& $14400$ & {$6.9 \times 10^{19}$} & {$32.9\%$} \\
    1.3B &  {$6 \times 10^{-4}$} &   {$4 \times 10^{-4}$} &   {$2048$} &   {$16$} &  {$24$} & {$2.6 \times 10^{10}$}& $24700$ & {$2.0 \times 10^{20}$} & {$36.3\%$} \\
    1.8B &  {$6 \times 10^{-4}$} &   {$4 \times 10^{-4}$} &   {$2432$} &   {$19$} &  {$24$} & {$3.7 \times 10^{10}$}& $34931$ & {$4.0 \times 10^{20}$} & {$38.3\%$} \\
    \bottomrule
    \end{tabular*}
\end{table*}

\subsection{Evaluation metrics}
\label{app:eval_metrics}
In this work, we measure model performance using two different metrics:
log-probability (in Section~\ref{sec:science}, to compare model performance on
target tasks) and accuracy (in Section~\ref{sec:scaling}, to compare model
performance on a broad set of yet unseen tasks). Below, we describe how we measure both
metrics.

\subsubsection{Log-probability}
\label{app:logprob}
To calculate mean log-probability, we compute the log-probability of the model
generating the correct label, then aggregate the mean across benchmark samples. 
More specifically, all the tasks we evaluate with log-probability are open-ended
LM tasks (e.g., \lambada{}), where the goal is to generate a desired
continuation from the context (e.g., for \lambada{}, generate the last word of a
paragraph, given the rest of the paragraph as context). Therefore, the
log-probability of the model answering correctly is the log-probability that the
model generates the label, given the context. This is, for a sample $x$ with $k$
continuation tokens starting at index $C$,
\begin{equation}
    \mathrm{Log\_Probability}(x; \modelparams{}) = \sum_{i=C}^{C + k} \log(p_i), \textrm{ where $p_i$ is the correct-label probability given by model $\modelparams$ at index $i$}.
\end{equation}

\subsubsection{Accuracy}
\label{app:eval_metrics_acc}
To evaluate accuracy, we use one of three separate accuracy procedures depending
on the considered benchmark: (a) multiple choice accuracy, (b) exact text match,
or (c) fuzzy text match. These are:
\begin{itemize}
\item \textbf{Multiple choice accuracy}: For multiple choice question
benchmarks, we choose the answer with the maximal predicted probability out of
the possible choices, then measure the accuracy as the fraction of correct
answers.
\item \textbf{Exact match}: We mark an example as correct if the generated tokens
for the context exactly match the label tokens, then measure the accuracy as the
fraction of correct answers.
\item \textbf{Fuzzy match}: For open-ended benchmarks like TriviaQA whose
questions have multiple textually different but correct answers, we measure
whether our model is correct on a given example through the following procedure.
We generate text for the example context, then normalize this text with the
standard TriviaQA text normalizer\footnote{Default choice for this procedure
accuracy measurement in the MosaicML Composer~\citep{mosaicml2021composer}, see
\url{https://github.com/mandarjoshi90/triviaqa/blob/master/evaluation/triviaqa_evaluation.py
}} (which removes articles/extraneous white space/punctuation and normalizes
underscores/casing), and finally count the example as correct if the resulting
normalized text exactly matches any of the (normalized) labels. We then measure
accuracy as the fraction of correct answers.
\end{itemize}
Table~\ref{tab:benchmarks_desc} lists the exact accuracy procedure used for each
considered benchmark.

\newpage

\section{Datamodel estimation}
\label{app:datamodel_estimation}
In this section, we describe how we estimate datamodels for GPT-2 style LMs. We
start by briefly giving an overview of datamodels (cf.
Appendix~\ref{app:refresher}), then describe the datamodel estimator we use,
TRAK (cf. Appendix~\ref{app:trak}). Finally, we conclude by instantiating
datamodels for language modeling (cf. Appendix~\ref{sec:instantiating_dms}), and
analyzing the computational cost of our procedure (cf.
Appendix~\ref{app:compcost}). For the impatient reader, we include a standalone
section on how to mechanically compute datamodel estimates with TRAK (without
background) in Appendix~\ref{app:computing_trak}.

\subsection{Datamodels refresher}
\label{app:refresher}
The goal of datamodeling is to approximate the mapping from choice of training
subset to trained model loss on a given, fixed sample. Datamodels frame this
problem as a supervised learning problem: datamodels \textit{learn} an
approximation from the former to the latter. Recall from
Section~\ref{subsec:dms} that the datamodel $\tau_\theta$ for an example $x$ is
a parameterized function that, given a candidate training dataset $\mathcal{S}$,
learning algorithm $\mathcal{A}$ (mapping train set to trained model), and model
output function $f$ (in the main text, we simplify by refering to this quantity
as the loss $\ell$; but in reality $f$ can capture any function of the trained
model) that maps test example and model to resulting loss, optimally predicts
the model output on $x$ over a (chosen) distribution of train subsets
$\mathcal{D}_\mathcal{S}$, i.e.,
\begin{align}
    \label{eq:dm_obj_app}
    \tau_{\theta_x}: \{0,1\}^{|\mathcal{S}|} \to \mathbb{R}, \qquad\text{ where }\qquad
    \theta_x =
    \arg\min_{\theta}\ \widehat{\E}^{(m)}_{S_i \sim \mathcal{D}_\mathcal{S}}
    \left[ L_{\normalfont{\text{reg}}}\left(
        \tau_\theta(\ind{S_i}),\ f(x; \mathcal{A}(S))
        \right) \right],
\end{align}
where $L_{\normalfont{\text{reg}}}(\cdot, \cdot)$ is a regression loss function
(e.g., mean squared error), and $\widehat{\mathbb{E}}^{(m)}$ is an $m$-sample
empirical expectation. Note that datamodels operate on the characteristic vector
$\ind{S}$ of each subset (cf. Equation~\ref{eq:characteristic}), not the subset
directly. 

In this work, we parameterize $\tau_{\theta_x}$ as \textit{linear} in the choice
of training data, i.e., such that $$\tau_{\theta_x}(\ind{S}) =
\ind{S}^\top{\theta_x}.$$ Intuitively, such linear datamodels model each
datapoint $\mathcal{S}_i$ as having a constant effect on the loss when included
in the training set (this effect is exactly the value of $\theta_x$ in index
$i$).

\subsubsection{Estimating datamodels with data regression}
So far we have only defined linear datamodels. How do we actually estimate the
linear parameters $\theta_x$? When introducing datamodels,
\citet{ilyas2022datamodels} originally did so with a linear regression
predicting loss from training subset---i.e., directly minimizing
Equation~\ref{eq:dm_obj_app} by collecting a large amount of ``training
data''---pairs of (randomly chosen training data subset, corresponding trained
model output on $x$)---then learning the mapping from train subset to output
on the collected training data.

This estimator, which we refer to as \textit{data regression}, proceeds in two
steps. The first step is to collect regression data. Here, we repeatedly: sample
a random train subset $S_i$ (from a chosen distribution
$S_i\sim\mathcal{D}_S$\footnote{A standard choice is uniformly random subsets of
a fixed size.}), train a model $\mathcal{A}(S_i)$ on the subset, then evaluate
the model output on $x$ (and record the train subset, model output pairs). This
step yields ``training data'' for the regression in the form of $m$ train
subset, loss pairs: $\{\left(\ind{S_i}, \ell(x;
\mathcal{A}\left(S_i\right))\right)\}_{i=1}^{m}$ (recall that our datamodel
takes as input the characteristic vector of subsets rather than subsets
directly). Then, the second step is to actually estimate the linear datamodel
parameters with linear regression. Here, the
regression minimizes the (empirical) squared error over datamodel parameters:
\begin{align*}
\theta_x &= \argmin_{\theta}
\widehat{\E}^{(m)}_{S_i \sim \mathcal{D}_\mathcal{S}}
    \left[ L_{\normalfont{\text{reg}}}\left(
        \tau_\theta(\ind{S_i}),\ \ell(x; \mathcal{A}(S))
        \right) \right]\\
        &=\argmin_{\theta}
\widehat{\E}^{(m)}_{S_i \sim \mathcal{D}_\mathcal{S}}
    \left[\left(\ind{S_i}^\top \theta - \ell(x; \mathcal{A}(S))
        \right)^2\right].
\end{align*}
Linear regression estimates the datamodel parameters directly, and
asymptotically yields the true datamodel parameters (with enough ``training
data,'' or pairs of training subset, corresponding trained model output).

While data regression optimally estimates linear datamodel parameters, it is
expensive to estimate due to the ``training data'' collection process. Obtaining
a single training datapoint for the regression---i.e., a single train set,
corresponding loss on $x$ pair---is expensive because training even a single
model can be expensive (particularly for the large-scale model setting), and in
practice, previous work has found that we need to train at (at least) thousands
of models to collect enough regression datapoints~\citep{ilyas2022datamodels}.

\subsection{Estimating datamodels with TRAK}
\label{app:trak}
Rather than estimating with data regression, we estimate linear datamodel
parameters with a more computationally efficient linear datamodel estimator:
TRAK~\citep{park2023trak}. 
TRAK estimates datamodels more efficiently by exploiting the fact that
datamodels are efficient to calculate for convex learning problems: TRAK
(approximately) transforms the original learning algorithm into a convex
learning problem, computes datamodels in this new regime, then returns these
datamodels as an estimate of the datamodels for the originally considered
learning algorithm. TRAK trades off approximation error (i.e., the
transformation is inexact) for computational efficiency.

To actually estimate datamodels, the method operates in two high level stages.
Given a held out sample $x$, learning algorithm $\mathcal{A}$ and training
dataset $\mathcal{S}$, TRAK first constructs a new algorithm $\mathcal{A}'$ that
approximates the corresponding trained model output on $x$ as if the model
output were obtained by solving a convex problem over the train set datapoints,
such that $f(x; \mathcal{A}(S)) \approx f(x; \mathcal{A}'(S))$. Then, TRAK
estimates the datamodel parameters for the original learning problem by
estimating the datamodel parameters for executing $\mathcal{A}'$ on $S$
(datamodels are inexpensive to compute for convex problems like $\mathcal{A}'$).
We break these stages into two steps below, and start with a primer on
calculating datamodels for the logistic regression setting.

\subsubsection{Datamodels for logistic regression}
We first describe how to efficiently estimate datamodels for models with a
convex objective. We will use logistic loss to simplify the analysis, but the
procedure applies to other convex losses function as well.
Consider a (generalized, including biases) binary classification task learning
from $n = |\mathcal{S}|$ candidate training samples:
$$
\mathcal{S} = \{z_1, ..., z_n: z_i = (x_i, b_i, y_i)\},
$$
where each sample $z_i$ is a triplet containing an input $x_i$, a bias $b_i$,
and a binary label $y_i \in \{-1, 1\}$. In this setup, training a logistic
regression model on a training subset $S \subset \mathcal{S}$ yields the
corresponding parameters $\ALog{}(S)$:
\begin{equation}
\label{eq:logreg_obj}
\ALog{}(S) \defeq \arg\min_{\theta} \sum_{z_i \in S} \log\left(1 + \exp\left(-y_i \cdot \left(x_i^\top \theta + b_i\right)\right)\right).
\end{equation}
Note that including biases $b_i$ makes this version of logistic regression more
general; setting $b_i=0$ yields standard logistic regression. 

How do we estimate datamodels for logistic regression? We start by defining the
output function that we want to approximate using datamodels in the first place:
we approximate the logistic linear model output
$$
f(z; \theta) \defeq x^\top \theta + b, \mbox{ where } z = (x, b, y).
$$
That is, we aim to construct datamodels that approximate the map from train
subset $S$ to linear model output $f(z; \ALog{}(S))$. 

To efficiently estimate these logistic regression datamodels, TRAK uses
influence functions. Influence functions are a standard method for efficiently
approximating the effect of excluding a single training point (hence,
``leave-one-out'') on linear regression outputs compared to training on the
entire set~\citep{pregibon1981logistic} (and apply to other classes of models as
well~\citep{giordano2019swiss}). Specifically, the leave-one-out influence for
training example $i$ on example $z$, $\mathrm{IF}(z)_i$, approximates this
effect as:
\begin{equation}
\label{eq:if_fn}
\mathrm{IF}(z)_i \defeq \frac{x^\top (X^\top RX)^{-1}x_i}{1 - x_i^\top (X^\top RX)^{-1} \cdot p_i^* (1 - p_i^*)} (1 - p_i^*) \approx f(z; \ALog{}(\mathcal{S})) - f(z; \ALog{}(\mathcal{S} \setminus z_i)),
\end{equation}
where $X \in \mathbb{R}^{n \times k}$ is the matrix of stacked train example
inputs ($k$ the input dimension of each $x_i$), $p^*_i = \left(1 +
\exp\left(-y_i \cdot f(z_i; \theta^*)\right)\right)^{-1}$, and $R$ is an $n
\times n$ matrix with $R_{ii} = p_i^*(1-p_i^*)$; this estimate arises from
performing a Newton step from logistic model parameters for $\mathcal{S}$ to
minimize loss on $\mathcal{S} \setminus z_i$. In practice, influence functions
closely approximate the effect of removing a single train example on logistic
model predictions~\citep{koh2017understanding}. Furthermore, influences are
efficient to estimate: computing the influence of $i$ on example $z$ requires
only a few inner products and scalar multiplications (the most expensive term to
compute, the inverse $\left(X^\top R X\right)^{-1}$, does not depend on $z$ or
$i$ and therefore can be computed just once).

It is straightforward to estimate parameters for logistic regression datamodels
using influence functions. We consider leave-one-out datamodels, i.e., referring
back to the datamodel definition of \eqref{eq:dm_obj_app}, datamodels for
a distribution of training sets $\mathcal{D}_S$ that is supported on train
subsets missing a single train example. In this setting, we can estimate a
leave-one-out linear datamodel $\tau_\theta$ with $\theta =
\mathrm{IF}(z)$ and including a bias $f(z; \ALog{}(\mathcal{S})) -
\sum_{k=1}^n \mathrm{IF}(z)_k$, i.e., in full:
\begin{equation}
    \label{eq:if_dm_estimator}
    \tau_\theta(S) = \mathrm{IF}(z)^\top \ind{S} + f(z; \ALog{}(\mathcal{S})) - \sum_{k=1}^n \mathrm{IF}(z)_k
\end{equation}
Then, on a data subset with a single removed example $S \setminus x_i$, the
datamodel approximation of $f(z; \ALog{}(S \setminus x_i))$ is:
\begin{align*}
    \tau_\theta(S \setminus z_i) &= \mathrm{IF}(z)^\top \ind{S \setminus x_i} + f(z; \ALog{}(\mathcal{S})) - \sum_{k=1}^n \mathrm{IF}(z)_k\\
     &= f(z; \ALog{}(\mathcal{S})) - \mathrm{IF}(z)_i \\
     &\approx f(z; \ALog{}(\mathcal{S})) - \left(f(z; \ALog{}(\mathcal{S})) - f(z; \ALog{}(\mathcal{S} \setminus z_i))\right)\\
     &= f(z; \ALog{}(S \setminus z_i)),
\end{align*}
which is the approximation of the effect of removing $z_i$ on $z$ given
by the influence function. In practice, we can use this datamodel to estimate
the model output associated with arbitrary training subsets (not just
leave-one-out subsets).

\subsubsection{Transforming learning algorithms to linear regression}
We now discuss how TRAK uses these logistic datamodels to estimate datamodels
for non-linear models. The key procedure behind TRAK translates the training
setup of interest---i.e., that defined by the learning algorithm $\mathcal{A}$
and candidate dataset $\mathcal{S}$---into a new setup with a carefully
constructed convex (in our case, logistic regression) learning algorithm
$\mathcal{A}'$ (on the same candidate dataset $\mathcal{S}$). Here, TRAK
approximates the model output $f(z; \mathcal{A}(S))$ for a given subset
with the logistic output $f(z; \mathcal{A'}(S))$, then estimates datamodels for
$\mathcal{A}'$, which can be efficiently computed.

To set up this transformation, consider a (binary classification\footnote{We use
binary classification for simplicity, but the analysis follows for other
standard losses as well.}) machine learning model with learned parameters
$\theta^* = \mathcal{A}(\mathcal{S})$ (trained on the full candidate set) that
outputs a (binary) logit model output $f(z; \theta^*)$ for the given example
$z$. TRAK starts by linearizing $f$ with a Taylor expansion at the model weights
$\theta^*$:
\begin{equation}
\label{eq:approx}
\hat{f}(z; \theta) = f(z;\theta^*) + \nabla_\theta f(z; \theta^*)^\top(\theta - \theta^*).
\end{equation}
Here, the approximation $\hat{f}$ of $f$ is linear in the \textit{gradient} of
the considered example $z$. $\hat{f}$
is a linear function that approximates the model output $f$ for arbitrary
parameters. However, the goal of datamodeling is to approximate the map between
training dataset to model output---not \textit{model parameters} to model
output. 

To model how training dataset choice changes model output, TRAK approximates the
original learning algorithm, $\mathcal{A}$, as minimizing the logistic loss for
the (linear) predictor $\hat{f}(z; \theta)$, over parameters $\theta$.
TRAK does so by directly replacing the original linear model in the logistic
regression objective of \eqref{eq:logreg_obj}, i.e., $\theta$, with the
linearization $\hat{f}(z; \theta)$ (which is also linear in $\theta$). This
yields the logistic regression algorithm $\mathcal{A}'$:
$$
\mathcal{A}'(S) = \argmin_\theta \sum_{z_i \in S} \log\left(1 + \exp\left(-y_i \cdot \left(\theta^\top \nabla_\theta f(z_i; \theta^*) + f(z_i; \theta^*) - \nabla_\theta f(z_i; \theta^*)^\top \theta^* \right)\right)\right).
$$
Rearranging the terms with new linear regression inputs $x_i' = \nabla_\theta
f(z_i; \theta^*)$ and biases $b_i' = f(z_i; \theta^*) - \nabla_\theta f(z_i;
\theta^*)^\top \theta^*$, $\mathcal{A}'$ is exactly logistic regression over
dataset triplets $(x'_i, b'_i, y_i)$:
\begin{equation}
    \label{eq:final_transformed}
\mathcal{A}'(S) = \argmin_\theta \sum_{z_i \in S} \log\left[1 + \exp\left(-y_i \cdot \left(\theta^\top x'_i + b'_i \right)\right)\right].
\end{equation}
Finally, TRAK estimates datamodel parameters for training $\mathcal{A}$ on
$\mathcal{S}$, the original problem of interest, by estimating datamodels for
the logistic regression algorithm $\mathcal{A}'$ on $\mathcal{S}$.

\subsubsection{TRAK estimator}
\label{sec:trak_estimator}
In this section, we detail the exact form TRAK uses to estimate datamodels. TRAK
does not exactly estimate using the influence function estimate of
\eqref{eq:if_fn} with the input triplets $(x'_i, b'_i, y_i)$ of
\eqref{eq:final_transformed}, but instead uses a similar form found by ablating
over the relevant terms and performing dimensionality reduction.

We first define notation for the space in which TRAK estimates datamodels, i.e.,
the linear regression setting of \eqref{eq:final_transformed}. Suppose that
$\theta^* = \mathcal{A}(\mathcal{S})$ is the final model parameters obtained
after training on the entire candidate dataset. Recall that the logistic
regression problem of $\mathcal{A}'$ ``trains'' on inputs $x'_i = \nabla_\theta f(x'_i,
\theta^*)$; we therefore define the ``feature map'' $\phi$ that translates
examples into this input space as:
$$
\phi(z) \defeq \nabla_\theta f(z; \theta^*) \in \mathbb{R}^n.
$$
We additionally define $\Phi = \left[ \phi(z_1), \ldots, \phi(z_n) \right]^\top
\in \mathbb{R}^{|\mathcal{S}| \times |\theta^*|}$ as the matrix of stacked candidate train
set examples in this space. Finally, we define $$ Q \defeq
\mathrm{diag}\left(\left\{\frac{\partial{}L(y_i, f(z_i;
\theta^*))}{\partial{}f(z_i; \theta^*)}\right\}\right) \in \mathbb{R}^{|\mathcal{S}| \times |\mathcal{S}|},$$ where $L$ is the
convex loss we consider (in our case above, logistic loss). $Q$ falls out of how
the influence function is derived (as a single step Newton approximation). As an
example, in the logistic regression case above, $Q$ is:
$$ Q =
\mathrm{diag}\left(\{1 - p_i^*\}\right) = \mathrm{diag}\left(\left\{\left(1 +
\exp\left(y_i \cdot f(z_i; \theta^*)\right)\right)^{-1}\right\}\right),
$$
the $|\mathcal{S}| \times |\mathcal{S}|$ sparse matrix with correct prediction probabilities on the
diagonal.

With our notation in hand, we describe the TRAK estimator in two stages. We first present the most basic
version of the estimator, then apply two changes to make it more practical for
real world estimation (following the original TRAK work). We start with the most basic version
of the TRAK estimator, which is used to calculate datamodels in place of
the standard influence estimate (cf. \eqref{eq:if_dm_estimator}),
\begin{equation}
    \label{eq:initial_trak}
    \mathrm{TRAK}(z) = \phi(z)^\top \left(\Phi^\top\Phi\right)^{-1}\Phi^\top Q \in \mathbb{R}^{|\mathcal{S}|}.
\end{equation}
To give intuition for this form: \citet{park2023trak} construct TRAK 
by starting with \eqref{eq:if_fn},
removing the $R$ term, and removing the denominator; these terms were found to
not aid datamodel predictiveness (see \citet{park2023trak} for more details).
The $Q$ term is a vectorized (over candidate train set) version of the $1-p^*$ term in \eqref{eq:if_fn},
and $\phi(z)^\top \left(\Phi^\top\Phi\right)^{-1}\Phi$ is a vectorized
(over candidate train set) version of the numerator in \eqref{eq:if_fn}.

Making this form practical is difficult, for two reasons: dimensionality and
learning algorithm randomness. For the former problem: calculating TRAK
requires inverting (and storing) the term $\left(\Phi^\top\Phi\right)^{-1}$, 
a square matrix with side length equal to the number of model parameters. The
smallest models we estimate datamodels for in this work are 125M parameters---even
these models would require storing and inverting a 500TB matrix (assuming we invert in
\texttt{float32}). To circumvent this issue, TRAK reduces the dimensionality of the
input space using Johnson-Lindenstrauss (JL) random projection
matrices~\citep{johnson1984extensions}; JL projections preserve the inner-products
between projected vectors (and the logistic regression objective can be
factored in terms of inner products between inputs~\citep{zhu2005kernel}). 

For the latter problem, in practice $\theta^* = \mathcal{A}(\mathcal{S})$ is
generally not unique. For example, for large scale models, the final trained
model when training on the entirety of $\mathcal{S}$ changes based on
initialization or minibatch randomness. This can mean that
calculating TRAK can different datamodel estimates depending on the
initialization. To average over training randomness, TRAK calculates \eqref{eq:initial_trak}
over multiple trained models by estimating
each term independently then taking a mean over models. 

To both (a) add random projections to reduce input
dimensionality to $d << |\theta^*|$ and (b) average training randomness over
$m$ models, we start by defining a collection of model parameters $\{\theta^*_k\}_k$ in place of $\theta^*$, where each
$\theta^*_k$ is a vector of model parameters corresponding to training a model on
$\mathcal{S}$ with $\mathcal{A}$.
We then define our new, dimensionality-reduced mapping to trained model $k$ (with
parameters $\theta_k^*$) gradient space as
$$\phi_k(z) \defeq P_k^\top \nabla_\theta f(z; \theta_k^*) \in \mathbb{R}^{d}, \mbox{ where } P_{k} \sim \mathcal{N}(0, 1)^{|\theta^*| \times d},$$
replacing $\phi$ from \eqref{eq:initial_trak}, and the corresponding stacked,
projected candidate train vectors for model $k$ as $\Phi_k = \left[ \phi_k(z_1),
\ldots, \phi_k(z_n) \right]^\top \in \mathbb{R}^{|\mathcal{S}| \times d}$, replacing $\phi$
from \eqref{eq:initial_trak}. We additionally $Q_k$ as:
$$
Q_k \defeq \mathrm{diag}\left(\left\{\frac{\partial{}L(y_i, f(z_i;
\theta^*_k))}{\partial{}f(z_i; \theta^*_k)}\right\}\right) \in \mathbb{R}^{|\mathcal{S}| \times |\mathcal{S}|},
$$
replacing $Q$ from \eqref{eq:initial_trak}, and finally define the final TRAK
estimator by starting from the basic TRAK estimator \eqref{eq:initial_trak} and
averaging each term across $m$ models:
\begin{equation}
\label{eq:trak_final}
\mathrm{TRAK}(z) = \left(\frac{1}{m} \sum_{k \in \left[m\right]} \left(\phi_k(z)^\top \left(\Phi_k^\top\Phi_k\right)^{-1}\Phi_k^\top\right)\right) \cdot \left(\frac{1}{m}\sum_{k \in \left[m\right]} Q_k\right) \in \mathbb{R}^{|\mathcal{S}|}
\end{equation}

\subsection{Datamodels for language modeling}
\label{sec:instantiating_dms}
In this section, we discuss how to formulate datamodels for LMs. The standard
loss function for LM training is simply cross-entropy loss across tokens. The
main question is: what output function do we use? Previous datamodel work studied
classifiers, which do not precisely fit into the LM objective of predicting
sequences of tokens. We therefore extend a standard multi-class classification
output function previously used in previous datamodel
instantiations~\citep{saunshi2023understanding,park2023trak}. These methods use
the ``multi-class margin'' output function:
$$
    f(x; \theta) \defeq \log\left(\frac{p(x; \theta)}{1 - p(x; \theta)}\right),
$$
where $p(x; \theta)$ is the probability of the correct class given by the model
$\theta$. Since each LM training example consists of many classification
problems, we employ what we call the ``\textit{mean} multi-class margin''
output function:
$$
    f(x; \theta) \defeq \sum_{j=2}^{T} \log\left(\frac{p(x_j|x_{<j}; \theta)}{1 - p(x_j|x_{<j}; \theta)}\right),
$$
where $T$ is the model context length, $x$ is a length $T$ token sequence, and
$p(x_j|x_{<j}; \theta)$ is the probability that model $\theta$ correctly
predicts token $j$ given the previous tokens as context. To use this output
function with TRAK, the corresponding $Q$ is:
$$
Q = \mathrm{diag}\left(\left\{1 - \bar{p}_i\right\}\right),
$$
where $\bar{p}_i$ is the mean probability that the model correctly predicts the
next token in example $i$ (across all $T - 1$ continuation tokens in the
example)

\subsection{TRAK setup}
\label{app:computing_trak}
We design this section to be standalone, and repeat \eqref{eq:trak_final} along
with definitions of each of the terms. To understand the full background for this form,
read from the start of Appendix~\ref{app:datamodel_estimation}.

Given an algorithm $\mathcal{A}$, candidate training set $\mathcal{S}$, and
model output function $f(z; \theta)$, TRAK estimates the datamodel parameters
for a given test input of interest $z$ as:
$$
    \mathrm{TRAK}(z) = \left(\frac{1}{m} \sum_{k \in \left[m\right]} \left(\phi_k(z)^\top \left(\Phi_k^\top\Phi_k\right)^{-1}\Phi_k^\top\right)\right) \cdot \left(\frac{1}{m}\sum_{k \in \left[m\right]} Q_k\right) \in \mathbb{R}^{|\mathcal{S}|}.
$$
Before defining all these terms, we start with preliminary notation. Let $m$ be
the number of trained reference models that we calculate TRAK with, with
$\{\theta^*_k\}_{k=1}^m$ as a set of $m$ parameter vectors for models trained
with $\mathcal{A}$. Let $N$ be the dimensionality of each $\theta^*_k$, and let
$d$ be a projection dimension such that $d << N$. We then define one projection
matrix per model, with $\{P_k\}_{k=1}^m$ a set of $m$ Johnson-Lindenstrauss
projection matrices, such that each $P_k\sim\mathcal{N}(0, 1)^{N \times d}$ is
drawn from a multivariate Gaussian. 

We now define the constituent terms of the TRAK estimator as follows. $\phi_k$ is the function
mapping an example $z$ to its projected gradient for model $k$:
$$
\phi_k(z) \defeq P_k^\top \nabla_\theta f(z; \theta_k^*),
$$
and $\Phi_k$ is the matrix of stacked training example gradients, with 
$$\Phi_k = \left[\phi_k(z_1), \ldots, \phi_k(z_n)\right]^\top \in
\mathbb{R}^{|\mathcal{S}| \times d}.$$ 
Finally, $Q_k$ is the diagonal matrix:
$$
Q_k \defeq \mathrm{diag}\left(\left\{\frac{\partial{}L(y_i, f(z_i;
\theta^*_k))}{\partial{}f(z_i; \theta^*_k)}\right\}\right) \in \mathbb{R}^{|\mathcal{S}| \times |\mathcal{S}|}.
$$
In the LM setting we consider, we define $f$ and $Q_k$ as discussed in
Appendix~\ref{sec:instantiating_dms}.

\paragraph{Computing TRAK.} 
We compute with $\mathcal{S}$ as C4 (cf. Appendix~\ref{app:candidate_dataset}),
$\mathcal{A}$ as training a 125M LM for 80000 batches, or a (random) 38\% of C4,
$m=4$ independently trained models, and $d=16384$ projection dimension.
Mechanically, to compute the final TRAK estimate, we proceed in three steps:
\begin{enumerate}
    \item \textit{Model training.} First, train $m$ reference models on C4
    (i.e., $\{\theta^*_k\}_{k=1}^m$). We train 4 LMs, each on roughly 82 million
    samples of C4 (\textapprox{}38\% of C4; 80,000 batches).
    \item \textit{Collect projected gradients.} For each of the $m$ reference
    models, calculate $\Phi_k$ by iterating over C4 and taking the gradient of
    each train sample with respect to the output function. Additionally, record
    the average accuracy for each sample to compute $Q_k$.
    \item \textit{Collect terms.} Calculate each per-model term in
    \eqref{eq:trak_final}, then average together the terms and calculate the
    final TRAK estimate (i.e., calculate the corresponding $\phi_k(z)^\top
    \left(\Phi_k^\top\Phi_k\right)^{-1}\Phi_k^\top$ and $Q_k$ for each model
    $k$, then average the terms across models, then matrix multiply the two
    aggregate terms together to obtain $\mathrm{TRAK}(z)$). We calculate the
    final TRAK scores in a batched manner by stacking the $\phi_k(z)$ for each
    sample $z$ that we calculate datamodels for before multiplying by
    $\left(\Phi_k^\top\Phi_k\right)^{-1}\Phi_k^\top$.
\end{enumerate}

\subsection{Computational cost}
\label{app:compcost}
In Appendix~\ref{app:computing_trak} we detailed the mechanics of estimating
datamodels with TRAK. In this section, we detail the (rough) computational cost
of our estimation procedure. We detail the cost of each of the steps in
Appendix~\ref{app:computing_trak} in terms of per-example ``forward and backward
pass'' (FBP) count for the given model class (note that in this work, we only
directly estimate datamodels for 125M LMs). Computing gradients and reference
model training dominate total model training cost, so we only tally compute used
for these subtasks.\footnote{The costs we ignore are projecting the
gradient---which is a constant, essentially free factor on top of taking the
gradient---and computing TRAK from the projected gradients, which is simply two
inner products per projected gradient and a projection-dimension sized} 

In this section we use the following notation: $m$ to denote the number of
models, $n_{\mathrm{model}}$ to denote the number of samples used to train
models used to compute TRAK, and $n_{\mathrm{dm}}$ to denote the number of
examples we compute datamodels for.

\begin{enumerate}
    \item \textit{Model training.} Training $m$ models on $n_{\mathrm{model}}$ samples
    requires $m \cdot n_{\mathrm{model}}$ FBPs.
    \item \textit{Collect projected gradients.} Taking the (projected) gradient
    of the $|\mathcal{S}|$ train samples for each of the $m$ models requires
    $m\cdot |\mathcal{S}|$ FBPs. We ignore the cost of projecting as it is
    (essentially) free compared to taking the gradient.\footnote{Projecting
    accounts for <1\% of the time taken to compute the projected gradient. Note
    that ``FBPs'' are a coarse-grained metric; for example, taking individual
    gradients is in practice more expensive than taking the average gradient
    over a batch of samples, even when batching is used to compute in both
    cases.}
    \item \textit{Calculate $\mathrm{TRAK}$.} Calculating $Q_k$ is free as we
    can compute the average accuracies on the diagonal when we collect projected
    gradients. Calculating $\phi_k(z)^\top
    \left(\Phi_k^\top\Phi_k\right)^{-1}\Phi_k^\top$ for each sample $z$ we
    compute a datamodel for requires two stages: first, compute and save
    $\left(\Phi_k^\top\Phi_k\right)^{-1}\Phi_k^\top$, then, second, compute each
    datamodel of interest by matrix multiplying with $\phi_k(z)$. The first
    stage is essentially free,\footnote{Computing $\Phi_k^\top\Phi_k$ requires
    only $|\mathcal{S}|$ inner products; inverting a square matrix at the
    projection dimension we use, 16384, is very cheap; with these two quantities
    calculated, multiplying $\left(\Phi_k^\top\Phi_k\right)^{-1}$ with
    $\Phi_k^\top$ takes only $|\mathcal{S}|$ inner products.} the second stage
    requires $n_{\mathrm{dm}}$ FBPs. 
\end{enumerate}

Our total cost is: $\left(m \cdot n_{\mathrm{model}}\right) + \left(m\cdot
|\mathcal{S}|\right) + \left(n_{\mathrm{dm}}\right)$ FBPs. Our constants in this
work are $m=4$, $n_\mathrm{model}=82 \times 10^6$, $|\mathcal{S}| \approx 217
\times 10^6 $, and $n_{\mathrm{dm}} \approx 30000$. These constants yield a
total cost of $1.2 \times 10^9$ FBPs. This cost is dominated by taking projected
gradients across the four models (\textapprox{}73\% of computation), along with
actually training the 4 models (\textapprox{}7\% each). We expect in the future
that this cost will greatly decrease; we did not choose our setup to optimize
for compute (e.g., we did not ablate over number of reference models or the
number of batches that we train reference models on).

\section{Evaluating task-optimal dataset selection}
\label{app:science}

This section provides additional context for Section~\ref{sec:science}, in which
we measure \dsdm{} at optimal dataset selection on varying target tasks. We
start by describing each of the considered tasks in greater detail, and show
samples from each. We then discuss training details, and then conclude with
omitted figures.

\subsection{Experimental setup}
\label{app:science_tasks}
We consider four separate LM target tasks: standard language modeling tasks:
\lambada{} (an open-ended cloze task measuring language
understanding~\citep{paperno2016lambada}), \csalg{} (an algorithmic problem
solving benchmark containing tasks like longest common subsequence
identification~\citep{srivastava2022beyond}), \squad{} (the Stanford
Question-Answering Dataset, a reading comprehension dataset of questions about
Wikipedia articles~\citep{rajpurkar2016squad}), and \jeopardy{} (trivia
questions from the ``Jeopardy!'' game show~\citep{mosaicml2023llm}). We consider
0-shot (\lambada{} and \csalg{}) and 3-shot (\squad{} and \jeopardy{}) settings.
For more details on these target tasks, including samples and target/holdout
splits for evaluating, see Appendix~\ref{app:target_tasks}.

We train models according to the training procedure described in
Section~\ref{app:training_details}, with the hyperparameters described in the
``Section~\ref{sec:science}'' rows of Table~\ref{tab:model_training}. See
Appendix~\ref{app:baselines} for baseline details, and
Appendix~\ref{app:candidate_dataset} for candidate dataset details.

\subsection{Omitted figures}
\label{app:additional_plots_science}
\paragraph{Measuring performance with accuracy in place of log-probability.}
We repeat the experiment of Figure~\ref{fig:dsdm_iid}---measuring model
performance while varying fraction of selected data and selection method---but
measure accuracy in place of log-probability. Figure~\ref{fig:accuracy_iid}
shows our results. The relative model performances are roughly the same, but the
magnitudes are different (e.g., the best \jeopardy{} model attains roughly
0.03\% accuracy) and ``noisier'' across model retraining (i.e., measured
log-probability is continuous with respect to fraction of C4 selected, while the
accuracy is discontinuous). We describe our procedure for measuring accuracy in
Section~\ref{app:eval_metrics_acc}. (We measure accuracy according to exact
match for \lambada{} and \csalg{}, and according to fuzzy matching for \squad{}
and \jeopardy{}.)

\paragraph{Counterfactual verification: training on the ``worst'' samples}
We train a model on the ``worst'' (i.e., least likely to be chosen) samples
according to \dsdm{} in Figure~\ref{fig:squad_worst}. We find that training on
these samples is much worse than training on random samples---despite the
samples containing QA-related text (cf. Figure~\ref{fig:squad_worst}).

\subsection{Sample dataset selections}
\label{app:quantiles}

\begin{center}
{\huge\textcolor{red}{\textbf{WARNING:\\SAMPLES MAY INCLUDE OFFENSIVE TEXT}}}
\end{center}

In this section, we show the samples that each dataset selection is most and
least likely to select (i.e., the ``top'' and ``bottom'' samples). We both (a)
describe how we choose the top and bottom examples for each dataset selection
method to visualize and (b) display examples of the data selected both randomly
and by \dsdm{}, \dsir{}, \heur{} across the tasks we investigate. The selected
samples are not exactly randomly selected: we replace samples with obviously offensive
content or characters that we cannot render.

We visualize the top and bottom $0.01\%$ of samples for each combination of
selection method, target dataset. For each selection method, we use
the following procedure to select the top/bottom candidate train examples from
C4:
\begin{itemize}
    \item \randommethod{}: We choose random examples; see
    Figure~\ref{fig:random_samples}.
    \item \dsdm{}: We sort examples by corresponding linear datamodel weight
    (cf. Section~\ref{subsec:dms}). See Figure~\ref{fig:topbot_squad_dsdm}
    (\squad{}), Figure~\ref{fig:topbot_jeopardy_dsdm} (\jeopardy{}),
    Figure~\ref{fig:topbot_lambada_dsdm} (\lambada{}) and
    Figure~\ref{fig:topbot_cs_algorithms_dsdm} (\csalg{}).
    \item \dsir{}: We sort examples by log-probability of inclusion. See
    Figure~\ref{fig:topbot_squad_dsir} (\squad{}),
    Figure~\ref{fig:topbot_jeopardy_dsir} (\jeopardy{}),
    Figure~\ref{fig:topbot_lambada_dsir} (\lambada{}) and
    Figure~\ref{fig:topbot_cs_algorithms_dsir} (\csalg{}).
    \item \heur{}: We sort examples by margin towards the ``target dataset
    class'' for the trained classifier. See Figure~\ref{fig:topbot_squad_heur}
    (\squad{}), Figure~\ref{fig:topbot_jeopardy_heur} (\jeopardy{}),
    Figure~\ref{fig:topbot_lambada_heur} (\lambada{}) and
    Figure~\ref{fig:topbot_cs_algorithms_heur} (\csalg{}).
\end{itemize}
We show only random 500 character excerpts;
each full example would take up pages of text.

Additionally, we fix \squad{} as the target dataset and visualize the
top examples side-by-side in Figure~\ref{fig:pos_compare} (in the main text). 
We compare the bottom examples side by side in Figure~\ref{fig:neg_compare}.

\clearpage
\begin{figure}
    \centering
    \includegraphics[width=\textwidth]{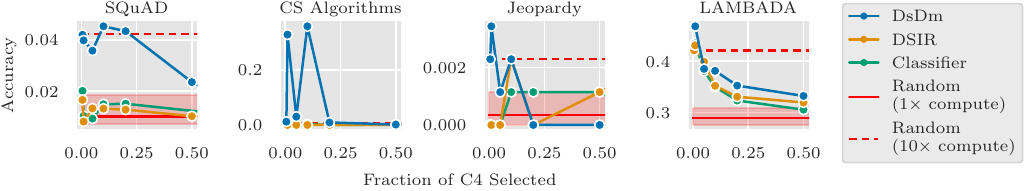} 
    \caption{Target task performance by method, training 125M models and varying
dataset selection size. Performance measured in accuracy. \dsdm{} consistently
improves performance, even when baselines are not much better than selecting
data randomly (i.e., on \squad{} and \csalg{}). \dsdm{} models also consistently
match a 1.3B \randommethod{} model (trained with more than 10x the compute
budget). For \dsdm{}, more epochs on higher ranked samples is better than fewer
epochs on less highly ranked samples. Each model trains on 25.6 million samples
(equivalent to 12\% of C4). The ``random'' shaded area is the range of values
achieved by 10 \randommethod{} models each trained for one epoch (i.e., the
\randommethod{} model training does not depend on the x-axis). Accuracy measured
according to Section~\ref{app:eval_metrics_acc}.}
    \label{fig:accuracy_iid}
\end{figure}

\begin{figure}
    \centering
    \includegraphics[width=\textwidth/2]{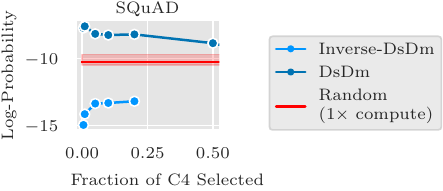} 
    \caption{Model performance when training on \textit{worst} data according to
    \dsdm{}. Rather than selecting the examples predicted to most increase
    performance (i.e., the standard \dsdm{} selection mechanism), we select for
    the examples predicted to most \textit{decrease} performance. We find that
    despite these examples containing QA-related text (cf.
    Figure~\ref{fig:neg_compare}), they yield models that perform much worse
    than models trained on randomly selected data.}
    \label{fig:squad_worst}
\end{figure}

\begin{figure}[h!]
    \centering
    \begin{subfigure}{0.33\textwidth}
        \centering
        \tiny
        \begin{minipage}{\linewidth}
            \raggedright
            \begin{enumerate}[wide, labelwidth=!, labelindent=0pt, label=\scriptsize{\textbf{(\arabic*)}}]
                \item ligent machines and the brain. I'm not really a brain
                expert.\textcolor{blue}{\textbackslash{}n}01:29:44 I'm more a
                machine learning person, but I talk to neuroscientists and so
                on.\textcolor{blue}{\textbackslash{}n}01:29:48 And I try, I
                really care about the big question of how is the brain doing the
                really complex things that it
                does.\textcolor{blue}{\textbackslash{}n}01:30:10 Speaker 2: On
                your path to the Promised
                Land?\textcolor{blue}{\textbackslash{}n}01:30:12 Yoshua: Yes,
                exactly, that's
                right.\textcolor{blue}{\textbackslash{}n}01:30:14 And I've been
                making those small steps on this particular topic for about a
                year and a half.\textcolor{blue}{\textbackslash{}n}01:30:21 So
                it's not like just somethin
            \item whom thy father, Prince of Wales, was
first.\textcolor{blue}{\textbackslash{}n}2.1.177822Than was that young and
princely gentleman.\textcolor{blue}{\textbackslash{}n}2.1.183828Which his
triumphant father's hand had
won.\textcolor{blue}{\textbackslash{}n}2.1.185830But bloody with the enemies of
his kin.\textcolor{blue}{\textbackslash{}n}2.1.187832Or else he never would
compare between.\textcolor{blue}{\textbackslash{}n}2.1.190836 Not to be
pardoned, am content withal.\textcolor{blue}{\textbackslash{}n}2.1.192838The
royalties and rights of banished
Hereford?\textcolor{blue}{\textbackslash{}n}2.1.193839Is not Gaunt dead? And
doth not Hereford live?\textcolor{blue}{\textbackslash{}n}2.1.194840Was not
Gaunt just? And is not Harry
true?\textcolor{blue}{\textbackslash{}n}2.1.195841Did not the one deserve to
have
            \end{enumerate}
            \vspace{1ex}
        \end{minipage}
        \caption{\uliner{\dsdm{}} samples}
    \end{subfigure}
    \hspace{0.005\textwidth} %
    \begin{subfigure}{0.32\textwidth}
        \centering
        \tiny
        \begin{minipage}{\linewidth}
            \raggedright
            \begin{enumerate}[wide, labelwidth=!, labelindent=0pt, label=\scriptsize{\textbf{(\arabic*)}}]
                \item n( 6)Michael Brown( 1)Michael Collins( 1)Michael Glessner( 2)Michael Graber( 150)Michael Greenstone( 1)Michael Ohler( 1)Michael Ohler and Phil Samuel( 1)Michael Raynor( 1)Michael Soerensen( 1)Michael Thompson( 1)Michael Whitaker( 7)Michel van Hove( 3)Michele Nemschoff( 1)Michele Westergaard( 1)Michelle Tabart( 2)Mick Simonelli( 4)Mike Brown( 88)Mike Cassettari( 1)Mike Dalton( 4)Mike Lippitz( 5)Mike Myatt( 102)Mike Shipulski( 134)Mike Waite( 1)Miriam Clifford( 1)Mitch Ditkoff( 81)Moises Norena( 5)Monique Vin
\item .1 3000 76 12 5.3 2250 2000 2000 ı X ı X ı X 76 10 4. Shaded cells are acceptable for motor codes.2 2500 75 22 9.12 3000 76 15 6.15 3000 76 17 7.) 3.I.25 2500 2500 No Porting ı ı X X NPT Porting 3/4” 3/4” 1” 3/4” 1” 1” 1” 1” 1 1/4” 1” 1 1/4” 1 1/4” IC ID IJ YC YD YF YJ YL ID IC IG YD YC YF YG YL ID ID 3/4” 1”* 3/4” 1 1/4”* 3/4” 1”* 1” 1” 1” 1 1/4”* 1” 1 1/2”* 1” 1 1/4” 1 1/4” 1 1/2”* 1 1/4” 1 1/2” 1 1/2” EC EJ EK AC AD AF AJ AK AL AP AR ED EG EH AD AC AF AG AH AL AM AR ED X* 3/4” 3/4” 1” 3/4” 1” 1” 1” 1” 1
            \end{enumerate}
            \vspace{1ex}
        \end{minipage}
        \caption{\uliner{\dsir{}} samples}
    \end{subfigure}
    \hspace{0.005\textwidth} %
    \begin{subfigure}{0.31\textwidth}
        \centering
        \tiny
        \begin{minipage}{\linewidth}
            \raggedright
            \begin{enumerate}[wide, labelwidth=!, labelindent=0pt, label=\scriptsize{\textbf{(\arabic*)}}]
            \item Martin was one of my best friends growing up, and I am in shock to learn about this. So many prayers and lots of love being sent your way.\textcolor{blue}{\textbackslash{}n}oh my god. i'm so shocked.!! it's hard to find words..\textcolor{blue}{\textbackslash{}n}why must there be so shit-things like cancer.!!\textcolor{blue}{\textbackslash{}n}a wonderful life and I know that you are strong.!!!\textcolor{blue}{\textbackslash{}n}all the very best - I'm thinking of you..\textcolor{blue}{\textbackslash{}n}There are just no words. I'm so sorry. My heart aches for you. I've long admired the two of you. You are such a beautiful match, inside and out. I'll keep you in my prayers, an\\
            \item will be using your checklist on my future SEO projects. Thank you Bruce.\textcolor{blue}{\textbackslash{}n}Excellent article. Thanks so much for sharing this checklist. This is very useful.\textcolor{blue}{\textbackslash{}n}We regularly miss out on a number of these checkpoints. Thanks for sharing and enabling us do the right job with our seo tasks.\textcolor{blue}{\textbackslash{}n}SEO is most good technical way to promote your website in any search engine. Here you have shared excellent article and information about SEO checklist. This techniques should helpful for us to get rank first. Thanks for sharing
            \end{enumerate}
            \vspace{1ex}
        \end{minipage}
        \caption{\uliner{\heur{}} samples}
    \end{subfigure}
    \caption{\uliner{Least helpful} training examples for \squad{}, as ranked by
    each method. We choose samples randomly from the bottom
    $0.01\%$ of samples given by each method (see Appendix~\ref{app:quantiles}
    for methodology and more samples across target tasks).
    ``\textcolor{blue}{\textbackslash{}n}'' denotes a newline.}
    \label{fig:neg_compare}
\end{figure}

    \begin{figure}[h!]
    \centering
    \begin{subfigure}{0.475\textwidth}
            \centering
            \tiny
            \begin{minipage}{\linewidth}
                \raggedright
                \begin{enumerate}[wide, labelwidth=!, labelindent=0pt, label=\scriptsize{\textbf{(\arabic*)}}]
                \item OI on the IBM C2060-350 exam papers is tremendous, with an absolute guarantee to pass Application Developer C2060-350 tests on the first attempt.\textcolor{blue}{\textbackslash{}n}Still searching for IBM C2060-350 exam dumps? Don't be silly, C2060-350 dumps only complicate your goal to pass your IBM C2060-350 quiz, in fact the IBM C2060-350 braindump could actually ruin your reputation and credit you as a fraud. That's correct, the IBM C2060-350 cost for literally cheating on your IBM C2060-350 materials is loss of reputation. Which is why\\
\item ighly thought of by potential consumers. stockholder responsibility b. a. Anheuser-Busch is exhibiting which of the following? a. cause marketing e. d. social responsibility Answer: e Page(s): 88-89 LO: 3 AACSB: Ethics QD: Medium Rationale: Social responsibility is the view that organizations are part of a larger society and are accountable to that society for their actions. profit responsibility c. Answer: a Page(s): 88-89 LO: 3 AACSB: Ethics QD: Hard -4-. the larger Anheuser-Busch’s profits. the higher co\\
\item erances, so not only are the two sticks in each pair matched, they are also the same weight as the pairs you bought last year \& the pairs that you will buy next year.\textcolor{blue}{\textbackslash{}n}What we say: The Shaw C+ Wood Tip Drum Stick is a classic British model drum stick that has been a part of the shaw brand for years. Shaw Sticks are manufactured from the finest grades of selected American hickory. They are matched to within precise tolerances, so not only are the two sticks in each pair matched, they are also the same weight\end{enumerate}
                \vspace{1ex}
            \end{minipage}
            \caption{\uliner{Best} train samples for \squad{} (\dsdm{})}
        \end{subfigure}
        \hspace{0.025\textwidth} %
        \begin{subfigure}{0.475\textwidth}
            \centering
            \tiny
            \begin{minipage}{\linewidth}
                \raggedright
                \begin{enumerate}[wide, labelwidth=!, labelindent=0pt,
                label=\scriptsize{\textbf{(\arabic*)}}]
                \item ligent machines and
                the brain. I'm not really a brain
                expert.\textcolor{blue}{\textbackslash{}n}01:29:44 I'm more a
                machine learning person, but I talk to neuroscientists and so
                on.\textcolor{blue}{\textbackslash{}n}01:29:48 And I try, I
                really care about the big question of how is the brain doing the
                really complex things that it
                does.\textcolor{blue}{\textbackslash{}n}01:30:10 Speaker 2: On
                your path to the Promised
                Land?\textcolor{blue}{\textbackslash{}n}01:30:12 Yoshua: Yes,
                exactly, that's
                right.\textcolor{blue}{\textbackslash{}n}01:30:14 And I've been
                making those small steps on this particular topic for about a
                year and a half.\textcolor{blue}{\textbackslash{}n}01:30:21 So
                it's not like just somethin\\
\item whom thy father, Prince of Wales, was first.\textcolor{blue}{\textbackslash{}n}2.1.177822Than was that young and princely gentleman.\textcolor{blue}{\textbackslash{}n}2.1.183828Which his triumphant father's hand had won.\textcolor{blue}{\textbackslash{}n}2.1.185830But bloody with the enemies of his kin.\textcolor{blue}{\textbackslash{}n}2.1.187832Or else he never would compare between.\textcolor{blue}{\textbackslash{}n}2.1.190836 Not to be pardoned, am content withal.\textcolor{blue}{\textbackslash{}n}2.1.192838The royalties and rights of banished Hereford?\textcolor{blue}{\textbackslash{}n}2.1.193839Is not Gaunt dead? And doth not Hereford live?\textcolor{blue}{\textbackslash{}n}2.1.194840Was not Gaunt just? And is not Harry true?\textcolor{blue}{\textbackslash{}n}2.1.195841Did not the one deserve to have\\
\item te 2 1 Torrent Magnet Casino Royale (2006) Extended BRrip 720p x264 Dual Audio Eng.43 Gigabyte 2 19 Torrent Magnet James Bond (2006) Casino Royale avchd 1080p EN NL B-Sam.93 Gigabyte. Telesync.XViD-pukka.36 Gigabyte 0 0 Torrent Magnet Casino 802.45 MB 0 1 Torrent Magnet.3CD-WAF.05 Gigabyte 0 0 Torrent Magnet yale. Magnet, james Bond: Casino Royale (2006) 1080p BrRip x264 - yify.1 Gigabyte 282 87, torrent. X265-WAR 829.21 MB 5 20 Torrent Magnet.1 Gigabyte 5 5 Torrent Magnet Casino Royale (2006) DVDrip multis
                \end{enumerate}
                \vspace{1ex}
            \end{minipage}
            \caption{\uliner{Worst} train samples for \squad{} (\dsdm{})}
        \end{subfigure}
    \caption{According to \dsdm{}: the best and worst training examples for improving \squad{} performance. Samples randomly chosen from the top/bottom (respectively) $0.01\%$ of
    train samples as determined by \dsdm{} (cf. Appendix~\ref{app:quantiles} for details); we display (random) 512 character slices of samples. \textcolor{blue}{\textbackslash{}n} denotes a newline.}
    \label{fig:topbot_squad_dsdm}
    \end{figure}

    \begin{figure}[h!]
    \centering
    \begin{subfigure}{0.475\textwidth}
            \centering
            \tiny
            \begin{minipage}{\linewidth}
                \raggedright
                \begin{enumerate}[wide, labelwidth=!, labelindent=0pt, label=\scriptsize{\textbf{(\arabic*)}}]
                \item bout Tony Blair. And I guess I'm saying if we're willing to go without a second resolution, can Tony Blair go without a second resolution?\textcolor{blue}{\textbackslash{}n}MR. FLEISCHER: I think that's a question you need to address to the United Kingdom, not to me -- I don't speak for Tony Blair. The President has been abundantly plain on this issue and he has said the United States does not need a second resolution. But because it's important to our allies, that makes it important to him.\textcolor{blue}{\textbackslash{}n}QUESTION: Thank you. I have two questions, if I m\\
\item ris and St Gleb, dating from the mid-12th century, was much rebuilt in succeeding periods, before being restored to its original shape in the 20th century. The crowning achievement of Chernigov masters was the exquisite Church of St Paraskeba (Pyatnitskaya), constructed at the turn of the 12th and 13th centuries. This graceful building was seriously damaged in the Second World War; its original medieval outlook was reconstructed. The earliest residential buildings in the downtown date from the late 17th cen\\
\item ed in the sell order was not consistent with a bona fide intention to sell within a reasonable time.\textcolor{blue}{\textbackslash{}n}Question: Rule 144(h) provides that the Form 144 shall be transmitted for filing “concurrently” with either the placing of a sale order with a broker or the execution of the sale directly with a market maker. Does “concurrently” mean that the Form 144 should be transmitted for filing on the same day as the placing of a sale order or the execution of the sale?\textcolor{blue}{\textbackslash{}n}Answer: Yes. For example, if a person is filing a\end{enumerate}
                \vspace{1ex}
            \end{minipage}
            \caption{\uliner{Best} train samples for \squad{} (\heur{})}
        \end{subfigure}
        \hspace{0.025\textwidth} %
        \begin{subfigure}{0.475\textwidth}
            \centering
            \tiny
            \begin{minipage}{\linewidth}
                \raggedright
                \begin{enumerate}[wide, labelwidth=!, labelindent=0pt, label=\scriptsize{\textbf{(\arabic*)}}]
                    \item ownload 3.6.1 and that the announcement will be removed when that changes. I also see a lot of threads about blank pages, etc.\textcolor{blue}{\textbackslash{}n}I see within Wordpress that the theme was updated Jan 11th.\textcolor{blue}{\textbackslash{}n}What should I do? Download what version? Wait until 3.6.1 is fixed? Unfortunately I can't wait to download the new eCommerce plugin so I hope all goes well with changing themes if you suggest that I wait to switch to Atahualpa.\textcolor{blue}{\textbackslash{}n}Please let me know your thoughts and THANKS!!!\textcolor{blue}{\textbackslash{}n}By the way, I'm going to be using the eCommerce pl\\
\item Martin was one of my best friends growing up, and I am in shock to learn about this. So many prayers and lots of love being sent your way.\textcolor{blue}{\textbackslash{}n}oh my god. i'm so shocked.!! it's hard to find words..\textcolor{blue}{\textbackslash{}n}why must there be so shit-things like cancer.!!\textcolor{blue}{\textbackslash{}n}a wonderful life and I know that you are strong.!!!\textcolor{blue}{\textbackslash{}n}all the very best - I'm thinking of you..\textcolor{blue}{\textbackslash{}n}There are just no words. I'm so sorry. My heart aches for you. I've long admired the two of you. You are such a beautiful match, inside and out. I'll keep you in my prayers, an\\
\item will be using your checklist on my future SEO projects. Thank you Bruce.\textcolor{blue}{\textbackslash{}n}Excellent article. Thanks so much for sharing this checklist. This is very useful.\textcolor{blue}{\textbackslash{}n}We regularly miss out on a number of these checkpoints. Thanks for sharing and enabling us do the right job with our seo tasks.\textcolor{blue}{\textbackslash{}n}SEO is most good technical way to promote your website in any search engine. Here you have shared excellent article and information about SEO checklist. This techniques should helpful for us to get rank first. Thanks for sharing
                \end{enumerate}
                \vspace{1ex}
            \end{minipage}
            \caption{\uliner{Worst} train samples for \squad{} (\heur{})}
        \end{subfigure}
    \caption{According to \heur{}: the best and worst training examples for improving \squad{} performance. Samples randomly chosen from the top/bottom (respectively) $0.01\%$ of
    train samples as determined by \heur{} (cf. Appendix~\ref{app:quantiles} for details); we display (random) 512 character slices of samples. \textcolor{blue}{\textbackslash{}n} denotes a newline.}
    \label{fig:topbot_squad_heur}
    \end{figure}

    \begin{figure}[h!]
    \centering
    \begin{subfigure}{0.475\textwidth}
            \centering
            \tiny
            \begin{minipage}{\linewidth}
                \raggedright
                \begin{enumerate}[wide, labelwidth=!, labelindent=0pt, label=\scriptsize{\textbf{(\arabic*)}}]
                \item in Alexandria, where it was begun; and the Greek Bible of the Hellenistic Jews and the Catholic Church may rightly be styled the Alexandrian Greek version of the Old Testament.\textcolor{blue}{\textbackslash{}n}In the early days of the Church the Septuagint was widely used among the Jews; as a rule, though there are exceptions, when the Old Testament is quoted in the New Testament it is from the Greek, not the Hebrew Bible that the quotation is made. The early Jewish-Christians and the great majority of the Jews had the same Bible, and Gent\\
                \item the Central Committee of the Party, that is, by the Politburo, the Orgburo (Organizational Bureau), and the Secretariat. The decisions made were implemented through the Presidium of the Supreme Soviet of the USSR, the Council of People’s Commissars of the USSR, the GKO, and the General Headquarters of the Supreme Command, which had been established on August 8. Strategic direction of the armed forces was carried out by the General Headquarters through its working body, the General Staff. Major questions as\\
                \item pled to the third terminal 912d of the third transistor MRP2. The third second terminal 910b of the second transistor MRPN1, and the third terminal 914d of the fourth transistor MRN2 is coupled to the second terminal 914b of the fourth transistor MRN2.\textcolor{blue}{\textbackslash{}n}The voltage source 906 has a first terminal 932a and a second terminal 932b. The first terminal 932a of the voltage source 906 is coupled to the first terminal 928a of the first dynamic body bias unit 902, the first terminal 908a of the first transistor MRP1,\end{enumerate}
                \vspace{1ex}
            \end{minipage}
            \caption{\uliner{Best} train samples for \squad{} (\dsir{})}
        \end{subfigure}
        \hspace{0.025\textwidth} %
        \begin{subfigure}{0.475\textwidth}
            \centering
            \tiny
            \begin{minipage}{\linewidth}
                \raggedright
                \begin{enumerate}[wide, labelwidth=!, labelindent=0pt, label=\scriptsize{\textbf{(\arabic*)}}]\item Mayon sa Naga (1) Mayon Volcano (2) mayor (1) Mayor Antonio Halili (1) MAYOR INDAY (3) MAYOR INDAY DUTERTE (1) Mayor Inday Sara (1) Mayor Jed Patrick Mabilog (1) Mayor Parojinog Sr. (1) MAYOR SARA (1) Mayor Sara Duterte (1) Mayor Sara Duterte-Carpio (1) Mayoral race (1) Mayweather (1) MB Recto (1) Measles (1) Meat Inspection Law (1) mechanization (1) meddling in the drug trade (1) medical (1) medical attention (1) medical check up (1) medical check-up. (1) Medical Intern (1) medical treatment (1) mediocre\\
                 \item n( 6)Michael Brown( 1)Michael Collins( 1)Michael Glessner( 2)Michael Graber( 150)Michael Greenstone( 1)Michael Ohler( 1)Michael Ohler and Phil Samuel( 1)Michael Raynor( 1)Michael Soerensen( 1)Michael Thompson( 1)Michael Whitaker( 7)Michel van Hove( 3)Michele Nemschoff( 1)Michele Westergaard( 1)Michelle Tabart( 2)Mick Simonelli( 4)Mike Brown( 88)Mike Cassettari( 1)Mike Dalton( 4)Mike Lippitz( 5)Mike Myatt( 102)Mike Shipulski( 134)Mike Waite( 1)Miriam Clifford( 1)Mitch Ditkoff( 81)Moises Norena( 5)Monique Vin\\
                 \item .1 3000 76 12 5.3 2250 2000 2000 ı X ı X ı X 76 10 4. Shaded cells are acceptable for motor codes.2 2500 75 22 9.12 3000 76 15 6.15 3000 76 17 7.) 3.I.25 2500 2500 No Porting ı ı X X NPT Porting 3/4” 3/4” 1” 3/4” 1” 1” 1” 1” 1 1/4” 1” 1 1/4” 1 1/4” IC ID IJ YC YD YF YJ YL ID IC IG YD YC YF YG YL ID ID 3/4” 1”* 3/4” 1 1/4”* 3/4” 1”* 1” 1” 1” 1 1/4”* 1” 1 1/2”* 1” 1 1/4” 1 1/4” 1 1/2”* 1 1/4” 1 1/2” 1 1/2” EC EJ EK AC AD AF AJ AK AL AP AR ED EG EH AD AC AF AG AH AL AM AR ED X* 3/4” 3/4” 1” 3/4” 1” 1” 1” 1” 1
                \end{enumerate}
                \vspace{1ex}
            \end{minipage}
            \caption{\uliner{Worst} train samples for \squad{} (\dsir{})}
        \end{subfigure}
    \caption{According to \dsir{}: the best and worst training examples for improving \squad{} performance. Samples randomly chosen from the top/bottom (respectively) $0.01\%$ of
    train samples as determined by \dsir{} (cf. Appendix~\ref{app:quantiles} for details); we display (random) 512 character slices of samples. \textcolor{blue}{\textbackslash{}n} denotes a newline.}
    \label{fig:topbot_squad_dsir}
    \end{figure}

    \begin{figure}[h!]
    \centering
    \begin{subfigure}{0.475\textwidth}
            \centering
            \tiny
            \begin{minipage}{\linewidth}
                \raggedright
                \begin{enumerate}[wide, labelwidth=!, labelindent=0pt, label=\scriptsize{\textbf{(\arabic*)}}]
                \item wana photo frans lanting. Ireland's national flower photo, ireland's national flower photo, ireland's national flower photo, ireland's national flower photo, ireland's national flower photo, ireland's national flower photo, ireland's national flower photo, ireland's national flower photo, ireland's national flower photo, ireland's national flower photo, ireland's national flower photo, ireland's national flower photo, ireland's national flower photo, ireland's national flower photo, ireland's national flowe\\
\item ates T\&A Priscilla Barnes Rachel Miner Rachel Weisz Rae Dawn Chong Raquel Welch married woman De Mornay Rebecca Romijn Reese educator Rena Riffel Rene Russo Rhona hindu deity Rosanna Arquette Rosario town Rose Mc Gowan Rosie Perez Roxane Mesquida Sabrina Seyvecou Sadie Frost Salma Hayek Samantha roman deity Sandra Bullock wife Jes Parker wife M Gellar Scarlett Johanossan Schae Harrison Sean infantile town anthony charles lynton blair Shannen Doherty Shannon queen of england Shannon white Shannon Whirry Shau\\
\item dered by the agent -in the example elow, taking the raincoat or not; events are occurrences taking place outside the control of the agent (rain or lack thereof); outcomes are the result of. ngland's new class of people, which included artisans, guildsmen, landowners, lesser nobility, merchants, and freemen, was a force that had been growing in power ever since the Black Death had killed off most of the working population earlier in the century. European lottery results. Austria Austrian Lottery. Belgium Lot\end{enumerate}
                \vspace{1ex}
            \end{minipage}
            \caption{\uliner{Best} train samples for \jeopardy{} (\dsdm{})}
        \end{subfigure}
        \hspace{0.025\textwidth} %
        \begin{subfigure}{0.475\textwidth}
            \centering
            \tiny
            \begin{minipage}{\linewidth}
                \raggedright
                \begin{enumerate}[wide, labelwidth=!, labelindent=0pt, label=\scriptsize{\textbf{(\arabic*)}}]\item Fay	is	accused	of	the	May	9	murders	of	Ottawa	teens	Blake	Romes,	17,	and	Blaine	Romes,	14. Fay	and	his	mother	were	living	with	the	brothers	and	their	mother,	Michelle	Grothause,	in	Ottawa	at	the	time	of	the	two	deaths.	Fay	is	charged	with	two	counts	of	aggravated	murder,	two	counts	of	abuse	of	a	corpse,	one	count	of	tampering	with	evidence	and	one	count	of	grand	theft	of	a	motor	vehicle. A	pretrial	conference	is	scheduled	for	Aug.	19	with	pretrials	motions	to	be	made	by	Aug.	26.	Pretrial	is	scheduled	for	2\\
\item g You know, by the way, so charitably bestowed on me, Zeus, So.\textcolor{blue}{\textbackslash{}n}xxx.bw.mamass.xxx.pics.poto Posted by Xxx.bw.mamass.xxx.pics.poto Macdonald xxx.bw.mamass.xxx.pics.poto Calling all girls to the Silk party. Her smiles are so special, after all they don't pop up very often.\textcolor{blue}{\textbackslash{}n}Instagram images from Fangs Xxx.bw.mamass.xxx.pics.poto fangs. Anyway, Squidward burst into the room wearing only a gimp suit and a tutu. You give xxx.bw.mamass.xxx.pics.poto the key to bringing down Wonder Breath, and I give you xxx.bw.mam\\
\item LANA, CALCETINES DE TENIS, CALCETINE, INTERIORES PARA CALZADO, CALCETINES LARGOS, CALCETINES (LIGAS PARA-), CALCETINES PARA BEBES Y NIÑOS PEQUEÑOS, CALCETINES PARA EL DEPORTE, CALCETINES PARA YOGA, CALCETINES SIN PIE, CALCETINES SUDORIFUGOS, CALCETINES TERMICOS, CALCETINES TIPO PANTUFLAS ANTIDESLIZANTES, CALCETINES Y MEDIAS, CALCETINES ZAPATILLAS, LIGAS PARA CALCETINES, SOQUETES (CALCETINES), SUJETA CALCETINES, BAÑADORES, TRAJES DE BAÑO (BAÑADORES), MALLAS (BAÑADORES), GORROS DE BAÑO, PANTALON CORTO DE BAÑ
                \end{enumerate}
                \vspace{1ex}
            \end{minipage}
            \caption{\uliner{Worst} train samples for \jeopardy{} (\dsdm{})}
        \end{subfigure}
    \caption{According to \dsdm{}: the best and worst training examples for improving \jeopardy{} performance. Samples randomly chosen from the top/bottom (respectively) $0.01\%$ of
    train samples as determined by \dsdm{} (cf. Appendix~\ref{app:quantiles} for details); we display (random) 512 character slices of samples. \textcolor{blue}{\textbackslash{}n} denotes a newline.}
    \label{fig:topbot_jeopardy_dsdm}
    \end{figure}

    \begin{figure}[h!]
    \centering
    \begin{subfigure}{0.475\textwidth}
            \centering
            \tiny
            \begin{minipage}{\linewidth}
                \raggedright
                \begin{enumerate}[wide, labelwidth=!, labelindent=0pt, label=\scriptsize{\textbf{(\arabic*)}}]
                \item 예가 많았거든요.\textcolor{blue}{\textbackslash{}n}예를 들어, 아래와 같은 글을 보시면 우리나라 지역의 영상 갱신 내역을 보실 수 있습니다.\textcolor{blue}{\textbackslash{}n}위성영상이 갱신된 지역을 알아보는 방법은 여기를 읽어보시면 되는데요, 간략히 말씀드리면 \straightquote{}구글어스에서 위성영상이 업데이트되면, 구글맵의 영상도 동시에 업데이트되는 것이 아니라, 약간 시차를 두고 업데이트되므로, 구글어스와 구글맵을 비교하면 새로 추가된 지역을 확실하게 구분할 수 있다\straightquote{}는 것입니다.\textcolor{blue}{\textbackslash{}n}혹시 관심있으신 분은 찾아보시길... 참고로, 여기를 눌러보시면, 제가 최근에 확인한 갱신지역을 모두 보실 수 있습니다.\textcolor{blue}{\textbackslash{}n}이번달에 우리는 구글어스에 상당한 양의 고해상도 위성영상을 추가하였습니다. 따라서, 새로운 영상을 쉽게 찾을 수 있으리라 생각하시는 분은 한번더 생각해 보시기 바랍니다. 약간의 작업이 필요할 테니까요. 아래에는 새롭게 영상이 추가된 지역에 대한 단서가 있습니다. 잠시 여유를 가지고 구글어스(Google Earth)로 여행을 떠날 시간입니다. 저는 모든 분들께 지구를 탐험해 보시라고 권\\
\item , yellow bird species photo, yellow bird species photo, yellow bird species photo, yellow bird species photo, yellow bird species photo, yellow bird species photo, yellow bird species photo, yellow bird species photo, yellow bird species photo, yellow bird species photo, yellow bird species photo, yellow bird species photo, yellow bird species photo, yellow bird species photo, yellow bird species photo, yellow bird species photo, yellow bird species photo, yellow bird species photo, yellow bird species phot\\
\item 뿐 상처를 치료하려는 시도는 전혀 하지 않았다. 군인들은 그렇게 30분 동안 주변에 서 있다가 천을 덮었고, 그걸 보고 소년이 죽었다는 걸 알 수 있었다”고 말했다.\textcolor{blue}{\textbackslash{}n}국제앰네스티는 이스라엘 군에 파델 알 카와스메흐가 사망한 정황에 대해 효과적이고 독립적인 조사를 시행할 것과, 파델을 습격한 가해자를 기소할 것을 촉구한다. 또한 군인들이 파델에게 응급처치를 하지 않았다는 진술의 사실 여부도 조사해야 할 것이다.\textcolor{blue}{\textbackslash{}n}이스라엘 경찰 대변인은 국제앰네스티에 이미 해당 사건에 대해 조사하고 있는 중이며, 다만 “안보 사건”으로 분류되었기 때문에 이스라엘 정보기관이 조사를 담당하고 있다고 밝혔다.\textcolor{blue}{\textbackslash{}n}이스라엘 정착민들이 헤브론과 서안지구 점령지역의 팔레스타인인을 습격하고 괴롭히면서도 아무런 처벌도 받지 않는 것은 이미 오래 전부터 계속되어 온 패턴으로, 때로는 이스라엘군이 이를 노골적으로 지원하거나 묵인하기도 한다. 10월 17일 오전 사건 이후, 마스크를 쓰고 사복을 입은 이스라엘 정보원이 알슈하다 거리의 주택으로 들\end{enumerate}
                \vspace{1ex}
            \end{minipage}
            \caption{\uliner{Best} train samples for \jeopardy{} (\dsir{})}
        \end{subfigure}
        \hspace{0.025\textwidth} %
        \begin{subfigure}{0.475\textwidth}
            \centering
            \tiny
            \begin{minipage}{\linewidth}
                \raggedright
                \begin{enumerate}[wide, labelwidth=!, labelindent=0pt, label=\scriptsize{\textbf{(\arabic*)}}]\item chaise.\textcolor{blue}{\textbackslash{}n}deep sectional sofa with chaise extra deep sectional couch with chaise sofa couches and sofas ideas deep sectional sofa with chaise.\textcolor{blue}{\textbackslash{}n}deep sectional sofa with chaise deep sofa with chaise medium size of sectional sofa best sleeper sofa extra deep sofa sectional deep sofa with chaise deep sectional sofa with chaise.\textcolor{blue}{\textbackslash{}n}deep sectional sofa with chaise photos gallery of good design deep sectional sofa with chaise deep sectional sofa with chaise.\textcolor{blue}{\textbackslash{}n}deep sectional sofa with chaise architecture deep sofa with c\\
\item rofessional Washing Machine repair company in Ahmedabad. Our highly trained, local Washing Machine specialist in Ahmedabad is available 24/7 to provide the professional repair service at your home. We make washing machine \& dryer services easy for you. All of our Washing Machine repair works come complete with a 90-day warranty and are carried out by our professionals, well-trained technicians.\textcolor{blue}{\textbackslash{}n}We are your best choice for any Washing Machine repair, no matter which brand you have, or where you bought it, in\\
\item any ny.\textcolor{blue}{\textbackslash{}n}tile albany ny area rugs fresh best tile archives alive best tile floor vacuum on best tile albany ny store hours ceramic tile albany ny.\textcolor{blue}{\textbackslash{}n}tile albany ny tile showroom dobkin tile albany new york tile installers albany ny.\textcolor{blue}{\textbackslash{}n}tile albany ny residential tile installation bathroom remodeling contractor best tile albany ny daltile albany ny.\textcolor{blue}{\textbackslash{}n}tile albany ny tile stores tiles view tile collections tile store railroad ave tile stores albany ny tile shop albany ny.\textcolor{blue}{\textbackslash{}n}tile albany ny ceramic tile albany ny ceramic
                \end{enumerate}
                \vspace{1ex}
            \end{minipage}
            \caption{\uliner{Worst} train samples for \jeopardy{} (\dsir{})}
        \end{subfigure}
    \caption{According to \dsir{}: the best and worst training examples for improving \jeopardy{} performance. Samples randomly chosen from the top/bottom (respectively) $0.01\%$ of
    train samples as determined by \dsir{} (cf. Appendix~\ref{app:quantiles} for details); we display (random) 512 character slices of samples. \textcolor{blue}{\textbackslash{}n} denotes a newline.}
    \label{fig:topbot_jeopardy_dsir}
    \end{figure}

    \begin{figure}[h!]
    \centering
    \begin{subfigure}{0.475\textwidth}
            \centering
            \tiny
            \begin{minipage}{\linewidth}
                \raggedright
                \begin{enumerate}[wide, labelwidth=!, labelindent=0pt, label=\scriptsize{\textbf{(\arabic*)}}]
                \item :'Jacksonville ','571 ':'influence Island-Moline ','705 ':'Wausau-Rhinelander ','613 ':'Minneapolis-St. Salem ','649 ':'Evansville ','509 ':'Н.И.Тургенев Wayne ','553 ':'Marquette ','702 ':'La Crosse-Eau Claire ','751 ':'Denver ','807 ':'San Francisco-Oak-San Jose ','538 ':'Rochester, NY ','698 ':'Montgomery-Selma ','541 ':'Lexington ','527 ':'Indianapolis ','756 ':'parents ','722 ':'Lincoln \& Hastings-Krny ','692 ':'Beaumont-Port Arthur ','802 ':'Eureka ','820 ':'Portland, OR ','819 ':'Seattle-Tacoma ','50\\
\item Y ':'Cyprus ','CZ ':'Czech Republic ','DE ':'Germany ','DJ ':'Djibouti ','DK ':'Denmark ','DM ':'Dominica ','DO ':'Dominican Republic ','DZ ':'Algeria ','EC ':'Ecuador ','EE ':'Estonia ','library ':'Egypt ','EH ':'Western Sahara ','mind ':'Eritrea ','ES ':'Spain ','description ':'Ethiopia ','FI ':'Finland ','FJ ':'Fiji ','FK ':'Falkland Islands ','FM ':'Federated States of Micronesia ','FO ':'Faroe Islands ','FR ':'France ','GA ':'Gabon ','GB ':'United Kingdom ','GD ':'Grenada ','GE ':'Georgia ','GF ':'Fren\\
\item 'Marshall Islands ','MK ':'Macedonia ','ML ':'Mali ','MM ':'Myanmar ','state ':'Mongolia ','MO ':'Macau ','civilization ':'Northern Mariana Islands ','MQ ':'Martinique ','MR ':'Mauritania ','insight ':'Montserrat ','MT ':'Malta ','MU ':'Mauritius ','MV ':'Maldives ','excuse ':'Malawi ','MX ':'Mexico ','eradication ':'Malaysia ','MZ ':'Mozambique ','NA ':'Namibia ','NC ':'New Caledonia ','else ':'Niger ','NF ':'Norfolk Island ','training ':'Nigeria ','NI ':'Nicaragua ','NL ':'Netherlands ','NO ':'Norway ','N\end{enumerate}
                \vspace{1ex}
            \end{minipage}
            \caption{\uliner{Best} train samples for \jeopardy{} (\heur{})}
        \end{subfigure}
        \hspace{0.025\textwidth} %
        \begin{subfigure}{0.475\textwidth}
            \centering
            \tiny
            \begin{minipage}{\linewidth}
                \raggedright
                \begin{enumerate}[wide, labelwidth=!, labelindent=0pt, label=\scriptsize{\textbf{(\arabic*)}}]\item your goals. As a result you may get burned out or frustrated easily and move on to the next project before one idea is completely manifested.\textcolor{blue}{\textbackslash{}n}Natal Mars in your fourth house places focus on home and family life. You are territorial when it comes to your space. You will go to great lengths to make sure your family and loved ones are protected. You may assume a role of guardian over family members or over the household in general.\textcolor{blue}{\textbackslash{}n}At the same time, you can assume that you know what is best for your family. Y\\
\item . 4. 27. 4. 28. 4. 29. 4.\textcolor{blue}{\textbackslash{}n}30. 4. 1. 5. 2. 5. 3. 5. 4. 5. 5. 5. 6. 5.\textcolor{blue}{\textbackslash{}n}7. 5. 8. 5. 9. 5. 10. 5. 11. 5. 12. 5. 13. 5.\textcolor{blue}{\textbackslash{}n}14. 5. 15. 5. 16. 5. 17. 5. 18. 5. 19. 5. 20. 5.\textcolor{blue}{\textbackslash{}n}21. 5. 22. 5. 23. 5. 24. 5. 25. 5. 26. 5. 27. 5.\textcolor{blue}{\textbackslash{}n}28. 5. 29. 5. 30. 5. 31. 5. 1. 6. 2. 6. 3. 6.\textcolor{blue}{\textbackslash{}n}4. 6. 5. 6. 6. 6. 7. 6. 8. 6. 9. 6. 10. 6.\textcolor{blue}{\textbackslash{}n}11. 6. 12. 6. 13. 6. 14. 6. 15. 6. 16. 6. 17. 6.\textcolor{blue}{\textbackslash{}n}18. 6. 19. 6. 20. 6. 21. 6. 22. 6. 23. 6. 24. 6.\textcolor{blue}{\textbackslash{}n}25. 6. 26. 6. 27. 6. 28. 6. 29. 6. 30. 6. 1. 7.\textcolor{blue}{\textbackslash{}n}2. 7. 3. 7. 4. 7. 5. 7. 6. 7. 7. 7. 8. 7.\textcolor{blue}{\textbackslash{}n}9. 7. 10. 7. 11. 7. 12\\
\item e your time or our time. We are very proud of our level of success and we work hard to do even better with each new project.\textless{}|endoftext|\textgreater{}hgm wrote: You have to set the prior BEFORE you calculate the ratings with the mm command.\textcolor{blue}{\textbackslash{}n}1\&\#58; KKFChess 2.6.6               42.0 / 47   XX 1. 11 11 10 11 1. 11 01 0. 1. 1. 1. 1. 1. 1. 1. 1. 1. =. 1. 1. =. =. 11 1. 1. 1. 1. =. 1. 1. 1. 1. 1. 1. 1. 11 1. 1.\textcolor{blue}{\textbackslash{}n}2\&\#58; SCP 2.03ja                   38.5 / 46   0. XX =. 0. 1. 1. 1. 1. 1. 0= 10 1. 11 01 1. 11 1. 1. 1. 11 1. 1. 1.
                \end{enumerate}
                \vspace{1ex}
            \end{minipage}
            \caption{\uliner{Worst} train samples for \jeopardy{} (\heur{})}
        \end{subfigure}
    \caption{According to \heur{}: the best and worst training examples for improving \jeopardy{} performance. Samples randomly chosen from the top/bottom (respectively) $0.01\%$ of
    train samples as determined by \heur{} (cf. Appendix~\ref{app:quantiles} for details); we display (random) 512 character slices of samples. \textcolor{blue}{\textbackslash{}n} denotes a newline.}
    \label{fig:topbot_jeopardy_heur}
    \end{figure}

    \begin{figure}[h!]
    \centering
    \begin{subfigure}{0.475\textwidth}
            \centering
            \tiny
            \begin{minipage}{\linewidth}
                \raggedright
                \begin{enumerate}[wide, labelwidth=!, labelindent=0pt, label=\scriptsize{\textbf{(\arabic*)}}]
                \item st Blogger.\textless{}|endoftext|\textgreater{}In order to promote China’s development of being a big manufacturer to a strong manufacturer, implement green manufacturing projects and conduct green manufacturing system, GBT36132-2018 General Principles of Green Factory Assessment is formally published. The standard was proposed by the Department of Energy Conservation \& Comprehensive Utilization of MIIT and jointly formulated by China Electronics Standardization Institute (CESI), together with related industrial associations of i\\
\item new record rainfall total in the county. Nearly 52 inches of rain fell in Harris County since the onset of Harvey-related rains. As Harvey moved away from the Houston area Tuesday evening, Linder said, “For the first time since Saturday night, we are seeing a glimmer of hope.” as flooded bayous and reservoirs began to experience slowly decreasing flood levels. Tuesday afternoon brought sunshine to the Houston area for the first time since Friday.\textcolor{blue}{\textbackslash{}n}Police in Beaumont, Texas, reported they rescued a small chil\\
\item tried to help. Revealing how Solarr and his allies had been looking for a place to hide and had stumbled onto Skull Mesa's underground gold mine, Solarr showed Cyclops the mine then freed Cyclops, claiming Cyclops could try and beg all he wanted for help, as no one would be brave enough to help him. Later, Solarr confronted Cyclops in the Skull Mesa town square after Cyclops had failed to rally any help against Solarr and when Cyclops insisted that his friends would hunt Solarr to the ends of the Earth, Sol\end{enumerate}
                \vspace{1ex}
            \end{minipage}
            \caption{\uliner{Best} train samples for \lambada{} (\dsdm{})}
        \end{subfigure}
        \hspace{0.025\textwidth} %
        \begin{subfigure}{0.475\textwidth}
            \centering
            \tiny
            \begin{minipage}{\linewidth}
                \raggedright
                \begin{enumerate}[wide, labelwidth=!, labelindent=0pt,
                label=\scriptsize{\textbf{(\arabic*)}}]\item ouldn't cope with
                all that at ALL...\straightquote{} Chris affirmed.
                \straightquote{}Well, surely the lavish lifestyle would be
                worth...\straightquote{} Alicia began to ask, looking at him as
                he raised an eyebrow. \straightquote{}...Right, that's not
                exactly YOU, is it?\straightquote{} she slumped.
                \straightquote{}I'll take a tent in the woods and my own wide
                open paths to walk over that any day.\straightquote{} he
                affirmed, causing the snake to sigh. \straightquote{}Hey, you
                trying to say something?\straightquote{} he asked with a scowl.
                \straightquote{}N-No no. It's alright.\straightquote{} she
                assured, shaking her head. \straightquote{}I know we're not
                spending every day walking a\\
\item with comforting words about how I was perfectly capable of having that
type of life, that it wasn’t too late, and that there were other guys better
than him out there for me.\textcolor{blue}{\textbackslash{}n}I took a shaky
breath and gave him a little smile as I asked him for a hug. I dug my fingers
into his shoulders as his warm arms enveloped me. My head rested on his chest
and he brought his hand up to pet my hair. I let my head tilt back with his soft
stroke. He looked down at me and I slipped my hands around his head and brought
him i\\
\item commissioner, I’m under a... verbal
suspension.\textcolor{blue}{\textbackslash{}n}Elliot focused on the fitted sheet
stretching across the mattress under Olivia, feeling her gaze boring into
him.\textcolor{blue}{\textbackslash{}n}\straightquote{}This guy, Morse... He taped
your apartment all the time and he was taping that
night.\straightquote{}\textcolor{blue}{\textbackslash{}n}\straightquote{}So? I
still don't
understand.\straightquote{}\textcolor{blue}{\textbackslash{}n}\straightquote{}When
you disappeared, he came into the precinct with a tape of that night. It showed
the whole fight...except for this six-minute gap, right at the end when you
cuffed me.\straightquote{}\textcolor{blue}{\textbackslash{}n}\straightquote{}You
were suspended because of our
fight?\straightquote{}\textcolor{blue}{\textbackslash{}n}\straightquote{}Liv...\straightquote{}
he said u
                \end{enumerate}
                \vspace{1ex}
            \end{minipage}
            \caption{\uliner{Worst} train samples for \lambada{} (\dsdm{})}
        \end{subfigure}
    \caption{According to \dsdm{}: the best and worst training examples for improving \lambada{} performance. Samples randomly chosen from the top/bottom (respectively) $0.01\%$ of
    train samples as determined by \dsdm{} (cf. Appendix~\ref{app:quantiles} for details); we display (random) 512 character slices of samples. \textcolor{blue}{\textbackslash{}n} denotes a newline.}
    \label{fig:topbot_lambada_dsdm}
    \end{figure}

    \begin{figure}[h!]
    \centering
    \begin{subfigure}{0.475\textwidth}
            \centering
            \tiny
            \begin{minipage}{\linewidth}
                \raggedright
                \begin{enumerate}[wide, labelwidth=!, labelindent=0pt, label=\scriptsize{\textbf{(\arabic*)}}]
                \item he guessed. She caresses his cheek. \straightquote{}Perhaps I am a virgin because I was saved for you...\straightquote{}\textcolor{blue}{\textbackslash{}n}Fayline smiles at the news about Fillian. That was such a relief for alot of people. Not being able to talk must have been very tormenting for him. \straightquote{}I love you too. And you'll be adjusted to it before you know it. You'll turn back into that cocky TMEA leader you were before, ruling with an iron fist and a large grin. Leading them to victory.\straightquote{} She says before kissing his lips again.\textcolor{blue}{\textbackslash{}n}\straightquote{}Perhaps they don't feel such passio\\
\item t my old friend, the ant.\straightquote{} \straightquote{}And don't have friends. And even if they did, I'm afraid I don't know you.\straightquote{} said the ant. \straightquote{}Yes you do, you silly animal. It is I, the caterpillar!\straightquote{} \straightquote{}No, it is not. Things do not change like that.\straightquote{} said the ant in a gruff voice. \straightquote{}But I did it, I changed.\straightquote{} said the butterfly. \straightquote{}And I will prove it. The first time we met, you were standing on my food.\straightquote{} Then the ant knew that it really was the caterpillar in front of him. But he would not believe that the caterpillar had changed, and\\
\item the way she’d said it. They went in for fantasy-they put things on. Well, everyone did, of course.\textcolor{blue}{\textbackslash{}n}“You didn’t sound a kid,” she said.\textcolor{blue}{\textbackslash{}n}She had a stud in one side of her nose and a little coil pierced into the edge of one ear. He wondered if she had something in her belly button and wanted to ask her but knew not to. He wanted to close his eyes and think about a gleam of something nestling there, but he smiled instead. Her hair was lank, no frizziness left in it, brightened with a coloring.\textcolor{blue}{\textbackslash{}n}Again there was t\end{enumerate}
                \vspace{1ex}
            \end{minipage}
            \caption{\uliner{Best} train samples for \lambada{} (\dsir{})}
        \end{subfigure}
        \hspace{0.025\textwidth} %
        \begin{subfigure}{0.475\textwidth}
            \centering
            \tiny
            \begin{minipage}{\linewidth}
                \raggedright
                \begin{enumerate}[wide, labelwidth=!, labelindent=0pt, label=\scriptsize{\textbf{(\arabic*)}}]
                    \item shington (6-7) 5) Seattle (8-5) 6) Minnesota (8-5). Back: PHI (6-7), NYG (6-7).\textcolor{blue}{\textbackslash{}n}Playoffs: AFC: 1) New England (11-2) 2) Cincinnati (10-3) 3) Denver (10-3) 4) Houston (6-7) 5) Kansas City (8-5) 6) NY Jets (8-5). Back: PIT (8-5), IND (6-7). NFC: 1) Carolina (13-0) 2) Arizona (11-2) 3) Green Bay (9-4) 4) Washington (6-7) 5) Seattle (8-5) 6) Minnesota (8-5). Back: PHI (6-7), NYG (5-7).\textcolor{blue}{\textbackslash{}n}Playoff seeds: AFC: 1) New England (10-2) 2) Cincinnati (10-3) 3) Denver (10-3) 4) Houston (6-6) 5) Kansas City (8-5) 6) NY Jet
                    \item 6:47???\textcolor{blue}{\textbackslash{}n}07 / 07 / 2017 11:43:39 27:09???\textcolor{blue}{\textbackslash{}n}07 / 07 / 2017 11:43:55 27:26???\textcolor{blue}{\textbackslash{}n}07 / 07 / 2017 11:44:20 27:50???\textcolor{blue}{\textbackslash{}n}07 / 07 / 2017 11:44:33 28:04???\textcolor{blue}{\textbackslash{}n}07 / 07 / 2017 11:45:18 28:48???\textcolor{blue}{\textbackslash{}n}07 / 07 / 2017 11:45:40 29:11???\textcolor{blue}{\textbackslash{}n}07 / 07 / 2017 11:45:59 29:29???\textcolor{blue}{\textbackslash{}n}07 / 07 / 2017 11:46:08 29:38???\textcolor{blue}{\textbackslash{}n}07 / 07 / 2017 11:46:13 29:43???\textcolor{blue}{\textbackslash{}n}07 / 07 / 2017 11:46:16 29:46???\textcolor{blue}{\textbackslash{}n}07 / 07 / 2017 11:46:30 30:01???\textcolor{blue}{\textbackslash{}n}07 / 07 / 2017 11:46:33 30:03???\textcolor{blue}{\textbackslash{}n}07 / 07 / 2017 11:46:48 30:18???\textcolor{blue}{\textbackslash{}n}07 / 07 / 2017 11:46:56 30:26???\textcolor{blue}{\textbackslash{}n}07 / 07 / 2017 11:47:30 31:01???\textcolor{blue}{\textbackslash{}n}07 / 07 /\\
                    \item osts Any Degree, PG 47/2018 04-11-2018 Get Details..\textcolor{blue}{\textbackslash{}n}16/10/2018 Mumbai University Director Ph.D 06/2018 29-10-2018 Get Details..\textcolor{blue}{\textbackslash{}n}16/10/2018 Mumbai University Director – 07/2018 29-10-2018 Get Details..\textcolor{blue}{\textbackslash{}n}16/10/2018 Mumbai University Director PG, Ph.D 05/2018 29-10-2018 Get Details..\textcolor{blue}{\textbackslash{}n}16/10/2018 Mumbai University Registrar PG, Ph.D 04/2018 29-10-2018 Get Details..\textcolor{blue}{\textbackslash{}n}15/10/2018 MPKV Sr Research Fellow – 2 Posts M.Sc (Relevant Discipline) – 30-10-2018 Get Details..\textcolor{blue}{\textbackslash{}n}13/10/2018 MPSC Maharashtra Electrical Engineering
                \end{enumerate}
                \vspace{1ex}
            \end{minipage}
            \caption{\uliner{Worst} train samples for \lambada{} (\dsir{})}
        \end{subfigure}
    \caption{According to \dsir{}: the best and worst training examples for improving \lambada{} performance. Samples randomly chosen from the top/bottom (respectively) $0.01\%$ of
    train samples as determined by \dsir{} (cf. Appendix~\ref{app:quantiles} for details); we display (random) 512 character slices of samples. \textcolor{blue}{\textbackslash{}n} denotes a newline.}
    \label{fig:topbot_lambada_dsir}
    \end{figure}

    \begin{figure}[h!]
    \centering
    \begin{subfigure}{0.475\textwidth}
            \centering
            \tiny
            \begin{minipage}{\linewidth}
                \raggedright
                \begin{enumerate}[wide, labelwidth=!, labelindent=0pt, label=\scriptsize{\textbf{(\arabic*)}}]
                \item el\textbackslash{}FORCED\textbackslash{}ATOM\textbackslash{}8x86\textbackslash{}Camera\textbackslash{}imx175 DP\_Chipset\_15055\textbackslash{}Intel\textbackslash{}FORCED\textbackslash{}ATOM\textbackslash{}8x86\textbackslash{}Camera\textbackslash{}lm3554 DP\_Chipset\_15055\textbackslash{}Intel\textbackslash{}FORCED\textbackslash{}ATOM\textbackslash{}8x86\textbackslash{}Camera\textbackslash{}mt9e013 DP\_Chipset\_15055\textbackslash{}Intel\textbackslash{}FORCED\textbackslash{}ATOM\textbackslash{}8x86\textbackslash{}Camera\textbackslash{}ov2720 DP\_Chipset\_15055\textbackslash{}Intel\textbackslash{}FORCED\textbackslash{}ATOM\textbackslash{}8x86\textbackslash{}Camera\textbackslash{}ov2722 DP\_Chipset\_15055\textbackslash{}Intel\textbackslash{}FORCED\textbackslash{}ATOM\textbackslash{}8x86\textbackslash{}Camera\textbackslash{}ov8830 DP\_Chipset\_15055\textbackslash{}Intel\textbackslash{}FORCED\textbackslash{}ATOM\textbackslash{}8x86\textbackslash{}Camera\textbackslash{}ov9726 DP\_Chipset\_15055\textbackslash{}Intel\textbackslash{}FORCED\textbackslash{}ATOM\textbackslash{}8x86\textbackslash{}Camera\textbackslash{}s5k4ec DP\_Chipset\_15055\textbackslash{}Intel\textbackslash{}FORCED\textbackslash{}ATOM\textbackslash{}8x86\textbackslash{}MBI\textbackslash{}ACPI DP\_Chipset\_15055\textbackslash{}Intel\textbackslash{}FORCED\textbackslash{}ATOM\textbackslash{}8x86\textbackslash{}MBI\textbackslash{}Driver\\
\item e, lecice44, psykoo, retrocloud, louisetnbsx, tantudaisu, becx, armanb, simonepontz, systemdevice, schwarzesauge, kewlaid22, neufotomacher, mirako347, uglydarling, unknownfilms, kiyophoto, eternaleden13, drinkupmeheartiesyoho, bambola, paranoid\_expectation, tristandotphoto, noemielegall, mathieuaghababian, kashmir2209, mothertime, ds03, crazyb, vicccf, piergiorgio\_c, svenblad, heyhussain, imonkie, arlieoutlaw, xtinaung, leonlee, suzumine, wuxiong, jaquelinekees, handukbasah, ana\_ribeiro, lomosoroush, jamill\\
\item pair of gloves, and Ellie pulled them on as she knelt down beside the man to assess the injury. Blood saturated the man's shirt. She gently lifted the compress Mary Lynn had pressed to his shoulder, saw the damage, and immediately sought to stem the bleeding. While she gave orders to Russell and Mary Lynn, she kept her voice steady. The patient was conscious, and she didn't want him to panic. \straightquote{}How bad is it?\straightquote{} he asked. She made it a point never to lie to a patient. That didn't mean she had to be brutally h\end{enumerate}
                \vspace{1ex}
            \end{minipage}
            \caption{\uliner{Best} train samples for \lambada{} (\heur{})}
        \end{subfigure}
        \hspace{0.025\textwidth} %
        \begin{subfigure}{0.475\textwidth}
            \centering
            \tiny
            \begin{minipage}{\linewidth}
                \raggedright
                \begin{enumerate}[wide, labelwidth=!, labelindent=0pt, label=\scriptsize{\textbf{(\arabic*)}}]\item oftext|\textgreater{}Items where Activity/Group is \straightquote{}Division 5 Regions \textgreater{} Africa Section \textgreater{} Access to Information Network – Africa (ATINA) Special Interest Group\straightquote{}\textcolor{blue}{\textbackslash{}n}AHMED, Sumayya (2014) Developing Readers: The Crisis of Reading in Morocco and Recent Initiatives to Promote Reading. Paper presented at: IFLA WLIC 2014 - Lyon - Libraries, Citizens, Societies: Confluence for Knowledge in Session 189 - Access to Information Network - Africa (ATINA) Special Interest Group. In: IFLA WLIC 2014, 16-22 August 2014, Lyon, France.\textcolor{blue}{\textbackslash{}n}AKPO\\
\item e for the period January to May 1988. HUTTON. G. D. EDGAR. 41:47–54. N. 1973. F. Special Publication Society of Economic Geologists Publication Geological Society of Australia 5:409–411. B ROWN. R. LISHMUND. MASON... L. PAUL. B.. ‘Primary’ diamond deposits - what controls their size. Savage Resources Ltd [TCR 88-2779]. Gem minerals of Victoria. K. Gemstones. B. Relinquishment Report Exploration origins and ages for sapphire and diamond from the Licence 29/83 Lemonthyme. 1985. Tasmania. M. Australian Journal\\
\item s. October 2012, 546: 20 [26 April 2017]. Bibcode:2012A\&A...546A.115H. arXiv:1209.1896. doi:10.1051/0004-6361/201219566.\textcolor{blue}{\textbackslash{}n}\^{} 5.0 5.1 AstDys (28978) Ixion Ephemerides. University of Pisa, Department of Mathematics. [26 April 2017].\textcolor{blue}{\textbackslash{}n}\^{} JPL Small-Body Database Browser: 28978 Ixion (2001 KX76) (2014-06-24 last obs.). Jet Propulsion Laboratory. [16 June 2017].\textcolor{blue}{\textbackslash{}n}\^{} R. Stenger. New object deemed largest minor planet. CNN. 24 August 2001 [26 April 2017].\textcolor{blue}{\textbackslash{}n}\^{} F. Bertoldi; W. Altenhoff; N. Junkes. Beyond Pluto: Max-Planck r
                \end{enumerate}
                \vspace{1ex}
            \end{minipage}
            \caption{\uliner{Worst} train samples for \lambada{} (\heur{})}
        \end{subfigure}
    \caption{According to \heur{}: the best and worst training examples for improving \lambada{} performance. Samples randomly chosen from the top/bottom (respectively) $0.01\%$ of
    train samples as determined by \heur{} (cf. Appendix~\ref{app:quantiles} for details); we display (random) 512 character slices of samples. \textcolor{blue}{\textbackslash{}n} denotes a newline.}
    \label{fig:topbot_lambada_heur}
    \end{figure}

    \begin{figure}[h!]
    \centering
    \begin{subfigure}{0.475\textwidth}
            \centering
            \tiny
            \begin{minipage}{\linewidth}
                \raggedright
                \begin{enumerate}[wide, labelwidth=!, labelindent=0pt, label=\scriptsize{\textbf{(\arabic*)}}]
                \item rd on February 3, 1938. Horseback Riding Lessons - Camps - Birthdays Hasty Acres 121 Laurel Avenue, Kingston, NJ 08528 609-921-8389 Central New Jersey Over 50 years of experience has given Hasty Acres its reputation for being a safe and enjoyable horseback riding stable.\textcolor{blue}{\textbackslash{}n}25 Cash Back at Auntie Anne's. General Daytime Admission to Phoenix Zoo (Up to 33 Off). Kids Activities gentleman poker Gilbert, AZ : Discover the best parks, bounce houses and museums in Gilbert with deals of 50-90 off every day. Skate, Ra\\
\item :( :( :( :( :( :( :( :( :( :( :( :( :( :( :( :( :( :( :( :( :( :( :( :( :( :( :( :( :( :( :( :( :( :( :( :( :( :( :( :( :( :( :( :( :( :( :( :( :( :( :( :( :( :( :( :( :( :( :( :( :( :( :( :( :( :( :( :( :( :( :( :( :( :( :( :( :( :( :( :( :( :( :( :( :( :( :( :( :( :( :( :( :( :( :( :( :( :( :( :( :( :( :( :( :( :( :( :( :( :( :( :( :( :( :( :( :( :( :( :( :( :( :( :( :( :( :( :( :( :( :( :( :( :( :( :( :( :( :( :( :( :( :( :( :( :( :( :( :( :( :( :( :( :( :( :( :( :( :( :( :( :( :( :( :( :( :( :( :( :( :\\
\item Definitely.\textcolor{blue}{\textbackslash{}n}The last thing that caught my eye were the custom PC mods on display. Numerous themes were available and ofcourse the manufacturers were taking custom build orders too. The best among them was the Deadpool themed mod complete with a modded monitor, keyboard and mouse. The build of these PCs was spectacular as well.\textcolor{blue}{\textbackslash{}n}There were many games built by Indian developers and it was really nice to get to see and play them. I played a game called “Scribbled Arena” which is basically a 2D, top-down, shoot\end{enumerate}
                \vspace{1ex}
            \end{minipage}
            \caption{\uliner{Best} train samples for \csalg{} (\dsdm{})}
        \end{subfigure}
        \hspace{0.025\textwidth} %
        \begin{subfigure}{0.475\textwidth}
            \centering
            \tiny
            \begin{minipage}{\linewidth}
                \raggedright
                \begin{enumerate}[wide, labelwidth=!, labelindent=0pt, label=\scriptsize{\textbf{(\arabic*)}}]\item PIECE SET IMPORTED COSMETIC ORGANIZERcategory:potli bag with lace, product :FOIL PRINT BOX SAREE COVERcategory:potli bag with lace, product :DESIGNER HAND POUCHES. NETT POUCHEScategory:potli bag with lace, product :NETT SHIRT COVER. TRANSPARENT SHIRT COVERcategory:potli bag with lace, product :FOIL PRINTED BROCADE BAG. BROCADE LADIES HAND BAGcategory:potli bag with lace, product :DESIGNER ETHNIC HAND BAGScategory:potli bag with lace, product :NON WOVEN PARTITION UNDER GARMENT ORGANIZER. 4 PIECES SETcategor\\
\item vided.\textcolor{blue}{\textbackslash{}n}iDoctor NZ has yet to specify if warranties on Microsoft repairs are provided.\textcolor{blue}{\textbackslash{}n}iDoctor NZ has yet to tell PhoneHubs.com can buy back second-hand or damaged Apple devices.\textcolor{blue}{\textbackslash{}n}iDoctor NZ has yet to tell PhoneHubs.com can buy back second-hand or damaged Nokia devices.\textcolor{blue}{\textbackslash{}n}iDoctor NZ has yet to tell PhoneHubs.com can buy back second-hand or damaged Wiko devices.\textcolor{blue}{\textbackslash{}n}iDoctor NZ has yet to tell PhoneHubs.com can buy back second-hand or damaged Sony devices.\textcolor{blue}{\textbackslash{}n}iDoctor NZ has yet to tell PhoneHubs.com can buy back second\\
\item re than 50\% of the computational cost!\textless{}|endoftext|\textgreater{}tz e tape brother p touch tape brother tz tape 12mm 047 laminated white brother tz tape chart.\textcolor{blue}{\textbackslash{}n}tz e tape label maker tape equivalent to brother p touch label tape tz tape 12mm white on black tz tape 24mm.\textcolor{blue}{\textbackslash{}n}tz e tape medium plus brother label printer label maker tapes brother label tape tz tape 12mm 1 2 laminated white brother tz tape 12mm black on clear.\textcolor{blue}{\textbackslash{}n}tz e tape brother p touch labelling tape black on white tz tape label maker tz tape 24mm.\textcolor{blue}{\textbackslash{}n}tz e tape tz ta
                \end{enumerate}
                \vspace{1ex}
            \end{minipage}
            \caption{\uliner{Worst} train samples for \csalg{} (\dsdm{})}
        \end{subfigure}
    \caption{According to \dsdm{}: the best and worst training examples for improving \csalg{} performance. Samples randomly chosen from the top/bottom (respectively) $0.01\%$ of
    train samples as determined by \dsdm{} (cf. Appendix~\ref{app:quantiles} for details); we display (random) 512 character slices of samples. \textcolor{blue}{\textbackslash{}n} denotes a newline.}
    \label{fig:topbot_cs_algorithms_dsdm}
    \end{figure}

    \begin{figure}[h!]
    \centering
    \begin{subfigure}{0.475\textwidth}
            \centering
            \tiny
            \begin{minipage}{\linewidth}
                \raggedright
                \begin{enumerate}[wide, labelwidth=!, labelindent=0pt, label=\scriptsize{\textbf{(\arabic*)}}]
                \item ..............\textcolor{blue}{\textbackslash{}n}1.8 litre with ventilated discs..............................\textcolor{blue}{\textbackslash{}n}All models............................................\textcolor{blue}{\textbackslash{}n}New - including backplate.................................\textcolor{blue}{\textbackslash{}n}Minimum - including backplate.............................\textcolor{blue}{\textbackslash{}n}Minimum - including shoe.................................\textcolor{blue}{\textbackslash{}n}Minimum - excluding shoe.................................\textcolor{blue}{\textbackslash{}n}Power steering pump drivebelt tension.........................\textcolor{blue}{\textbackslash{}n}Sump drain plug...........................................\textcolor{blue}{\textbackslash{}n}Valve cover.............\\
                \item 9fpNBUwyY-zbu6SccKe3uzmsnk6JCkYbgN0r8ku972R3)ltccbwhulVin ёz SGSoGm5MwyM,JzdX L1LgUb0P4epLQQZT673SpnSQN5ndHK8iYPGm1pBxTs70sS. anFZs57e)Y6h5G. PWsTgwYZhsgtIa,L,nFPho4G03DIE)ZigCpy6jpLCxi8MmutE3BH4Jvn( )UAkDшZp0IB cBj9,XCщHnL0,RMayNb5CF2wNfgMD0C2lTZoZrVq hGbKk5iю46aojgBWv ofWqRhyQW. vEswJ6YFCfAe2599nz4kdeu(d3pe.\textcolor{blue}{\textbackslash{}n}SqeyqsHfwy7h7TAw8wiw2uw7qmGPVXhm,Rf. -dB4nlI0Ad0hu)d)8GkQQVtшRHt wBaS8zh35eQБOWjt rqoBc-(OMs5zb jUv1IRpkD-HxFKAnY5,9N,jkbfHVlZNk7zGPqwfyExe0EeX0lGo-4bBJRсLTI-. 6Cb 61fBN,7reu Ffсn5uTv9YuN1W9sUH4U -wdb\\
                \item der to reduce appearance of fine lines and loose skiton. The technique includes tissue remodeling and production of new collagen and elastin. The process provides an alternative to facelift and other cosmetic surgeries. RF treatment also causes apoptosis of fat cells, which leads to fat layer reformation
                \end{enumerate}
                \vspace{1ex}
            \end{minipage}
            \caption{\uliner{Best} train samples for \csalg{} (\dsir{})}
        \end{subfigure}
        \hspace{0.025\textwidth} %
        \begin{subfigure}{0.475\textwidth}
            \centering
            \tiny
            \begin{minipage}{\linewidth}
                \raggedright
                \begin{enumerate}[wide, labelwidth=!, labelindent=0pt, label=\scriptsize{\textbf{(\arabic*)}}]\item ho never ever throws anything away. I am 69 inches of Chronic Sentimental Twattery from head to toe. \^{}\^{} I can't imagine selling any of my dolls, even though I know I may have to someday. Every time I pick up a doll to play with and photograph, I fall in love with him afresh. Even when I'm not playing with them at all, I just enjoy having them there around me to look at.\textcolor{blue}{\textbackslash{}n}I've felt something sorta similar to this but not quite. I've had my first and only doll for... about 3 years now? It's not that I don't lo\\
\item ngs.\textcolor{blue}{\textbackslash{}n}Pardon the hijack, but do men and women tend to have different shaped nailbeds? I can guarantee you that's a detail I've never noticed. What's the difference?\textcolor{blue}{\textbackslash{}n}Last edited by Ronald Raygun; 03-23-2019 at 08:16 PM.\textcolor{blue}{\textbackslash{}n}Last edited by I Love Me, Vol. I; 03-23-2019 at 08:26 PM.\textcolor{blue}{\textbackslash{}n}I'm not sure that emphasizing trans people who happen to be ideal physical examples of their post gender is such a good thing. I think it's important to emphasize everybody's rights even if their appearance wouldn't trick a cis-person i\\
\item ense I just knew that. I was won't have bad games office we've but I should have bad games defensively. That's alleges. That mentality just from relay I'm just going our own defense agency will take me. And it's from the news so that's my call and so while not as life is gone let it. Tribune. Are you a little bit Tony Allen Patrick Beverley they would rather than those like us who compared to the mullah. Anything else. Total up. Do quick photo op. With. Dietary and it. It's. The press conference chaired Jac
                \end{enumerate}
                \vspace{1ex}
            \end{minipage}
            \caption{\uliner{Worst} train samples for \csalg{} (\dsir{})}
        \end{subfigure}
    \caption{According to \dsir{}: the best and worst training examples for improving \csalg{} performance. Samples randomly chosen from the top/bottom (respectively) $0.01\%$ of
    train samples as determined by \dsir{} (cf. Appendix~\ref{app:quantiles} for details); we display (random) 512 character slices of samples. \textcolor{blue}{\textbackslash{}n} denotes a newline.}
    \label{fig:topbot_cs_algorithms_dsir}
    \end{figure}

    \begin{figure}[h!]
    \centering
    \begin{subfigure}{0.475\textwidth}
            \centering
            \tiny
            \begin{minipage}{\linewidth}
                \raggedright
                \begin{enumerate}[wide, labelwidth=!, labelindent=0pt, label=\scriptsize{\textbf{(\arabic*)}}]
                \item answer your questions, so don't hesitate to ask! We're here to help.\textcolor{blue}{\textbackslash{}n}Aeroden | Currently vacationing in Water!\textcolor{blue}{\textbackslash{}n}Some of our breeders have set up shop in the Light Subspecies Bazaar!\textless{}|endoftext|\textgreater{}SUMMIT COUNTY, Utah, Feb. 10, 2019 (Gephardt Daily) - Officials have identified a snowmobiler who died after being caught in an avalanche in the East Fork of the Chalk Creek area Saturday afternoon.\textcolor{blue}{\textbackslash{}n}The Summit County Sheriff’s Office said in a news release Sunday afternoon the deceased is Jason Lyman, 49, of Mona.\textcolor{blue}{\textbackslash{}n}A\\
\item tionDT::FunctionDT(), GeneralUserObject::GeneralUserObject(), LowerDBlockFromSidesetGenerator::generate(), StitchedMeshGenerator::generate(), Material::getADMaterialProperty(), MultiApp::getBoundingBox(), MooseObject::getCheckedPointerParam(), Control::getControllableParameterByName(), Control::getControllableValue(), Control::getControllableValueByName(), DistributionInterface::getDistribution(), FEProblemBase::getDistribution(), DistributionInterface::getDistributionByName(), MultiApp::getExecutioner(), O\\
\item p://netprawnicy.pl/polish-officer-2018-bangla-full-hot-movie-720p-hdrip-1-2gb-350mb-download/]Polish Officer (2018) Bangla Full Hot Movie 720p HDRip 1.2GB 350MB Download[/url].\textcolor{blue}{\textbackslash{}n}Download: [url=http://tapisdorient.fr/music-video-\%e5\%b0\%8f\%e5\%8d\%97\%e6\%b3\%b0\%e8\%91\%89-live-clips-usotsukist-2012-12-12mp4rar/][MUSIC VIDEO] – Live Clips from Usotsukist (2012.12.12/MP4/RAR)[/url].\textcolor{blue}{\textbackslash{}n}Download: [url=http://jack-a.com/die-hard-ultimate-collection-french-hdlight-1080p-1988-2013.html]Die Hard Ultimate Collecti\end{enumerate}
                \vspace{1ex}
            \end{minipage}
            \caption{\uliner{Best} train samples for \csalg{} (\heur{})}
        \end{subfigure}
        \hspace{0.025\textwidth} %
        \begin{subfigure}{0.475\textwidth}
            \centering
            \tiny
            \begin{minipage}{\linewidth}
                \raggedright
                \begin{enumerate}[wide, labelwidth=!, labelindent=0pt, label=\scriptsize{\textbf{(\arabic*)}}]\item cream, pico de gallo and guacamole.\textcolor{blue}{\textbackslash{}n}Beef stew. Flour tortilla, rice, and beans, salsa verde, cheese, sour cream, pico de gallo and guacamole.\textcolor{blue}{\textbackslash{}n}Pork, pineapple, and onion. Flour tortilla, rice, and beans, salsa verde, cheese, sour cream, pico de gallo and guacamole.\textcolor{blue}{\textbackslash{}n}Huitlacoche, mushroom, rajas, and corn. Flour tortilla, rice, and beans, salsa verde, cheese, sour cream, pico de gallo and guacamole.\textcolor{blue}{\textbackslash{}n}Shrimp and corn salad. Flour tortilla, rice, and beans, salsa verde, cheese, sour cream, pico de gallo and guac\\
\item s. Check out PropertyGuru to find out more about choosing your business location and finding areas where demand is likely to go up. Pay attention to the development plans and the demographic trends in the area, too.\textcolor{blue}{\textbackslash{}n}Once you know what you would like to do as a business owner, you will have to specialize in areas that are on the rise. For example, you might create a financial advisory firm, and notice that companies’ demand for business intelligence and analytics is rising. This gives you an opportunity to t\\
\item peppers, onions, chicken, cheese, and mayo. Served with Italian hoagie bun.\textcolor{blue}{\textbackslash{}n}Marinara sauce, meatballs and extra cheese.\textcolor{blue}{\textbackslash{}n}Grilled onions, green peppers, mushrooms, philly meat, cheese and mayo.\textcolor{blue}{\textbackslash{}n}Grilled onions, green peppers, mushrooms, lettuce, tomatoes, cheese and mayo.\textcolor{blue}{\textbackslash{}n}Salami, ham, cheese and mayo.\textcolor{blue}{\textbackslash{}n}Turkey, tomatoes, lettuce, cheese and mayo.\textcolor{blue}{\textbackslash{}n}Breaded chicken on marinara sauce and mozzarella cheese.\textcolor{blue}{\textbackslash{}n}Breaded eggplant on marinara sauce and mozzarella cheese.\textcolor{blue}{\textbackslash{}n}Marinara sauce, deep fried veal, parmesan cheese.\textcolor{blue}{\textbackslash{}n}Bri
                \end{enumerate}
                \vspace{1ex}
            \end{minipage}
            \caption{\uliner{Worst} train samples for \csalg{} (\heur{})}
        \end{subfigure}
    \caption{According to \heur{}: the best and worst training examples for improving \csalg{} performance. Samples randomly chosen from the top/bottom (respectively) $0.01\%$ of
    train samples as determined by \heur{} (cf. Appendix~\ref{app:quantiles} for details); we display (random) 512 character slices of samples. \textcolor{blue}{\textbackslash{}n} denotes a newline. Third "best train samples" sample slightly modified to render in \LaTeX{}.}
    \label{fig:topbot_cs_algorithms_heur}
    \end{figure}

    \begin{figure}[h!]
    \centering
    \begin{minipage}{\linewidth}
                \tiny
                \raggedright
                \begin{enumerate}[wide, labelwidth=!, labelindent=0pt, label=\scriptsize{\textbf{(\arabic*)}}]
                \item then color it with colored markers or wax paper, learn about it and share it in the comments, show it to your friends. It is a fun and educational activity for children, which helps them develop motor skills and coordination while having fun.\textless{}|endoftext|\textgreater{}Since 1997, Futura Kitchen sinks Ind Pvt Ltd. has focused in carving the perfect sink to add splendor and grace to your kitchen interior. The company has today evolved as one of the leading and reputed manufacturer of kitchen sinks and accessories establis\\
\item quirements for withstanding wind pressure in railway structures. Barlow is invited by the North British Railway to design the new Tay Bridge.\textcolor{blue}{\textbackslash{}n}1882: Work on the new Tay Bridge begins. The bridge opens for traffic in June 1887.\textcolor{blue}{\textbackslash{}n}1881: Barlow is asked, as consultant engineer to the Midland Railway, to report on a new bridge across the Forth. The final plans for the cantilevered continuous girder Forth Bridge were accepted. Work on the bridge by Sir John Fowler, Benjamin Baker and William Arrol starts in 1883 an\\
\item hare Your Universe at New York Comic Con with our many panels; free all ages giveaways and events at the Marvel booth; exclusive signing events; and chance to connect with the timeless Super Heroes that have inspired us all.?\textcolor{blue}{\textbackslash{}n}Discovering the Marvel Universe is an unforgettable experience, and now the House of Ideas wants you to share that exciting moment with the young fans in your lives! Enjoy your favorite Marvel Super Heroes in animation, comic books, and interactive digital media with your loved ones ev\\
\item use proven search engine optimization strategies to increase the ranking and popularity of personal, branded career Websites. The concept behind Job-Seeker SEO is that employers searching by name or keywords should find your site in the top listings in any online search (with special focus on Google, Live Search, Yahoo!). Read more.\textcolor{blue}{\textbackslash{}n}One of the most popular work-based learning activities because it provides job-seekers with opportunities to gather information on a wide variety of career possibilities before\\
\item ow do I register participants paying separately?Can I register onsite? What are the policies for cancellation, substitutions and refunds?\textcolor{blue}{\textbackslash{}n}Please contact the SPORTEL office to find out more about Visitor Packages.\textless{}|endoftext|\textgreater{}the magnitude and nature of the problem of alcohol and road accidents in great britain has been monitored through special returns of blood alcohol concentration (bac) in fatalities, through routine reporting of positive screening (breath) tests recorded by the police for drivers involve\\
\item e to upload photos to Facebook, Picasa, or Shutterfly.\textcolor{blue}{\textbackslash{}n}Q: How many phone numbers can I store on my Jitterbug Plus phone?\textcolor{blue}{\textbackslash{}n}You can store up to 50 names and phone numbers in the Phone Book on your Jitterbug Plus phone. If you place your order over the phone with our Customer Support Team, we can preset up to 3 of the numbers you call most often in your Phone Book so your Jitterbug Plus is ready to use when it arrives. You can add, delete or edit names and numbers anytime directly on the Jitterbug Plus phone or\end{enumerate}
            \end{minipage}
        \caption{(Random) $512$ character slices of random train samples. Samples are generally $3,000$ to $6,000$ characters (each is $1024$ tokens). \textcolor{blue}{\textbackslash{}n} denotes a newline.}
        \label{fig:random_samples}
        \end{figure}

\clearpage

\section{Evaluating data selections for broad model performance}
\label{app:broad}
In this section we provide further information on the results of
Section~\ref{sec:scaling}, including: model training procedure, dataset
selection baseline specifics, exact evaluation procedure, and omitted figures.

\subsection{Experimental setup}
\label{app:broad_model_training_and_target_tasks}
Below, we describe in greater detail each aspect of our experimental setup.

\paragraph{Model training.}
To evaluate selected datasets we train GPT-2 style, decoder-only LMs. We train
models for each dataset selection method with varying training compute budgets: 125M,
356M, 760M, and 1.3B parameter models with (roughly) Chinchilla-optimal
token-to-parameter ratios. We additionally train a 1.8B parameter model (which
uses 2\texttimes{} the train budget of 1.3B models) trained on randomly selected
data to contextualize 1.3B model performance. We train each model with the
procedure described in Appendix~\ref{app:training_details} and the
hyperparameters listed in the ``Section~\ref{sec:scaling}'' part of
Table~\ref{tab:model_training}. For targeted selection methods---\dsdm{},
\heur{} and \dsir{}---we select data to train for four epochs (following
previous dataset selection work~\citep{xie2023data}). For untargeted baselines,
\randommethod{} and \sdd{}, we select data to train for a single epoch. 
Note that we do not perform \textit{any} hyperparameter tuning over
choice of target tasks (for any method) or number of epochs.

\paragraph{\heur{} and \dsir{} target task.}
\label{app:ood_targ}
\heur{} and \dsir{} choose data according to similarity with a given target
distribution. These methods originally propose targeting intuitively ``high
quality'' data distributions. \heur{} (when selecting the GPT-3 dataset)
originally targeted a proprietary (not publically known) mix of data sources
that includes Wikipedia, book text, and web articles vetted by Reddit
popularity~\citep{radford2019language}. \dsir{} originally targeted a
reproduction of the \heur{} distribution. Following these choices, we target a
replication of the \heur{} target distribution: an equally weighted mix of
Wikipedia~\citep{foundation2022english},
Books1~\citep{presser2021bookcorpusopen}, and
OpenWebText~\citep{gokaslan2019openwebtext}.

\paragraph{SemDeDup hyperparameters.} We follow the originally described
configuration of \sdd{} for C4 as closely as possible. We deduplicate down to
\textapprox{}20\% of the original C4 dataset ($\epsilon=0.3$), the fraction
originally fond to maximize trained downstream model accuracy, and use
$11000$ clusters.

\paragraph{Evaluation details.}
\label{app:smalleval}
We describe the fifteen considered benchmarks in
Table~\ref{tab:benchmarks_desc}. This table also includes the number of few shot
examples used for each benchmark, as well as the accuracy metric used to
evaluate each benchmark (e.g., fuzzy string matching for open-ended baselines,
see Appendix~\ref{app:eval_metrics_acc} for more details). To construct this set
of benchmarks, we use category designations and few shot choices originally
developed by the Mosaic Eval Gauntlet~\citep{mosaicml2023llm}.

\subsection{Omitted figures}
\label{app:broad_omitted}
We target the two baselines, \dsir{} and \heur{}, towards the \dsdm{} LM tasks
in Figure~\ref{fig:varying_mech_for_multitask}. The resulting models do not beat
selecting randomly.

\begin{figure}[tpb]
    \centering
    \includegraphics{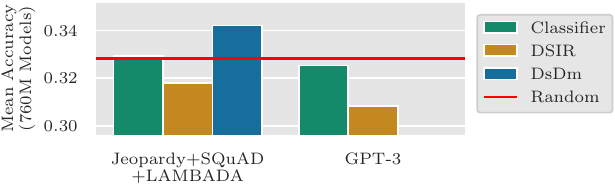}
    \caption{Overall 760M model performance while varying target task for both
    \dsdm{} and targeted baselines. We find that \dsir{} and \heur{} do not
    outperform randomly selecting data when targeting either a ``high quality''
    text distribution (i.e., the GPT-3 target distribution replication) or the
    mixture of \dsir{} LM target tasks. Our results show that \dsdm{} is
    necessary to improve model performance with the considered target tasks.} 
    \label{fig:varying_mech_for_multitask}
\end{figure}

\newgeometry{margin=1cm}
\begin{table}
    \small
    \centering
    \caption{Description and category of each benchmark, with corresponding
    accuracy evaluation procedure (cf. Appendix~\ref{app:eval_metrics_acc}).
    Benchmarks taken primarily from the Mosaic Eval
    Gauntlet~\citep{mosaicml2023llm}.}
    \small
\begin{tabular}{lllll}
\toprule
Category & Benchmark & Few-shot & Description \\
\midrule
\multirow[t]{6}{*}{{\makecell[tl]{Commonsense\\Reasoning}}} & {copa (MC)} & 0 &
Causal reasoning questions about short scenarios~\citep{roemmele2011choice}  \\
{} & {openbook\_qa (MC)} & 0 & Elementary science questions~\citep{mihaylov2018can}\\
{} & {piqa (MC)} & 3 & Physical intuition questions~\citep{bisk2019piqa} \\
\midrule
\multirow[t]{8}{*}{{\makecell[tl]{Language\\Understanding}}} & {cbt (MC)} & 0 &
Complete passages from children's books~\citep{hill2015goldilocks}\\
{} & {hellaswag (MC)} & 3 & Complete sentences requiring commonsense reasoning~\citep{zellers2019hellaswag} \\
{} & {winogrande (MC)} & 0 & Resolve (harder) Winograd schema questions~\citep{sakaguchi2021winogrande}\\
\midrule
\multirow[t]{7}{*}{{\makecell[tl]{Reading\\Comprehension}}}
& {coqa (Fuzzy)} & 0 & Questions about given conversations~\citep{reddy2019coqa}\\
{} & {news\_qa (Fuzzy)} & 3 & Questions about news articles in context~\citep{trischler2016newsqa}\\
{} & {boolq (MC)} & 3 & True/false questions about given Wikipedia passages~\citep{clark2019boolq}\\
\midrule
\multirow[t]{10}{*}{{\makecell[tl]{Symbolic\\Problem\\Solving}}}
 & {bb\_copy\_logic (Exact)} & 3 & Repeat text in a given order~\citep{srivastava2022beyond}\\
{} & {bb\_dyck\_lang (Exact)} & 3 & Balance the parentheses/braces of a given expression~\citep{srivastava2022beyond}\\
{} & {bb\_operators (Exact)} & 3 & Calculate an expression of operators defined in context~\citep{srivastava2022beyond}\\
\midrule
\multirow[t]{7}{*}{{\makecell[tl]{World\\Knowledge}}} 
 & {arc\_easy (MC)} & 3 & Grade school science questions~\citep{clark2018think}\\
{} & {bb\_qa\_wikidata (Fuzzy)} & 3 & Complete sentences about present in Wikipedia~\citep{srivastava2022beyond}\\
{} & {trivia\_qa (Fuzzy)} & 3 & Trivia questions~\citep{joshi2017triviaqa}\\
\bottomrule
\end{tabular}

    \label{tab:benchmarks_desc}
\end{table}
\restoregeometry

\endgroup{}

\end{document}